\documentclass[acmsmall]{acmart}

\usepackage{amsmath}
\usepackage{amsthm}
\usepackage{algorithm}
\usepackage{algpseudocode}
\usepackage{url}
\PassOptionsToPackage{hyphens}{url}
\usepackage{hyperref}
\usepackage{makecell}
\usepackage{multirow}
\usepackage{graphicx}
\usepackage{wrapfig}
\graphicspath{ {images/} }
\usepackage{color}

\usepackage[caption=false,font=footnotesize,labelfont=sf,textfont=sf]{subfig}

\newcommand{\dataset}{{\mathcal D}}

\algnewcommand{\LineComment}[1]{\State \(\triangleright\) #1}
\newcommand{\redtext}[1]{\textcolor{red}{#1}}
\newcommand{\bluetext}[1]{\textcolor{blue}{#1}}

\AtBeginDocument{%
  \providecommand\BibTeX{{%
    \normalfont B\kern-0.5em{\scshape i\kern-0.25em b}\kern-0.8em\TeX}}}

\setcopyright{acmcopyright}
\copyrightyear{2022}
\acmYear{2022}
\acmDOI{}

\acmJournal{TIST}
\acmVolume{37}
\acmNumber{4}
\acmArticle{111}
\acmMonth{10}

\begin{document}

\title{Selecting and Composing Learning Rate Policies for Deep Neural Networks}

\author{Yanzhao Wu}
\email{yanzhaowu@gatech.edu}
\orcid{0000-0001-8761-5486}
\author{Ling Liu}
\email{lingliu@cc.gatech.edu}
\affiliation{%
  \institution{Georgia Institute of Technology}
  \streetaddress{266 Ferst Drive}
  \city{Atlanta}
  \state{Georgia}
  \country{USA}
  \postcode{30332-0765}
}


\begin{abstract}
The choice of learning rate (LR) functions and policies has evolved from a simple fixed LR to the decaying LR and the cyclic LR, aiming to improve the accuracy and reduce the training time of Deep Neural Networks (DNNs). This paper presents a systematic approach to selecting and composing an LR policy for effective DNN training to meet desired target accuracy and reduce training time within the pre-defined training iterations. It makes three original contributions. First, we develop an LR tuning mechanism for auto-verification of a given LR policy with respect to the desired accuracy goal under the pre-defined training time constraint. Second, we develop an LR policy recommendation system (LRBench) to select and compose good LR policies from the same and/or different LR functions through dynamic tuning, and avoid bad choices, for a given learning task, DNN model and dataset. Third, we extend LRBench by supporting different DNN optimizers and show the significant mutual impact of different LR policies and different optimizers. Evaluated using popular benchmark datasets and different DNN models (LeNet, CNN3, ResNet), we show that our approach can effectively deliver high DNN test accuracy, outperform the existing recommended default LR policies, and reduce the DNN training time by 1.6$\sim$6.7$\times$ to meet a targeted model accuracy.
\end{abstract}


\maketitle

\section{Introduction} \label{sec:introduction}
Hyperparameter tuning is widely recognized as a critical optimization for efficient  training of deep neural networks (DNNs). A deep neural network is trained iteratively over the input training data $\dataset$ through forward and backward processes to update a set of trainable model parameters $\Theta$ based on the configuration of its  hyper-parameters $\mathcal{H}$ and the optimization algorithm of its loss function. The learning rate ($\eta$) is one of the most important hyper-parameters for efficiency optimization of the DNN training algorithms. The learning rate (LR) function and the configuration policy are known to have direct impacts on both the training efficacy and the test accuracy of the trained model. However, it is challenging to choose a good LR function, to select a good LR policy (e.g., a specific LR parameter configuration) given an LR function, and avoid bad LR policies. Even for the fixed learning rate function, it is non-trivial to choose a good value and avoid a bad one, since too small or too large LR value may impair the DNN training progress on both model accuracy and training time, resulting in slow convergence or even model divergence~\cite{Bengio-PracticalRecommendations,clr-3}.
The typical trial-and-error approach will try different LR values each time for training, which is tedious and time-consuming to tune the single LR value. Even with a reduced search space such as $[0.0001, 0.1]$, the possible LR values for trial-and-error can be inexplicable.
Bearing with the difficulty of determining a good LR value for fixed learning rate, a growing trend of research efforts have been devoted to more complex LR functions ($\eta(t;\mathcal{P})$), which have multiple LR parameters instead of a single fixed value, and will change as a function of the training iterations ($t$). As a result, finding a good LR function and selecting a good LR policy will demand tuning multiple LR parameters for each LR function, making the hyperparameter tuning for LR a far-reaching challenge~\cite{random-search,clr,sgdr,LRBenchBigData,autolr,autolrs}. 
Moreover, good LR policies for a given LR function tend to vary based on the specific datasets and learning tasks, and the DNN algorithms used for model training~\cite{optuna,dawnbench,ray-tune,GTDLBenchICDCS,GTDLBenchBigData,GTDLBenchTSC}. 
In practice, empirical approaches are typically used to manually select an LR function and configure a good LR policy by choosing the concrete LR parameters through trials and errors. For example, most of the DL frameworks (e.g., TensorFlow, Caffe and PyTorch) recommend different LR policies for different benchmark learning tasks and datasets as their default LR policies in their public releases with accuracy/training time benchmark results. Even for the same learning task and dataset, each of these DL frameworks often has different learning rate policies for different DNN models as the recommended default LRs. For example, TensorFlow uses a constant learning rate (fixed LR) with a specific LR value as its recommended LR policy for CIFAR-10 when training using AlexNet, and uses a decaying LR (NSTEP) as its default LR policy when ResNet is used for training a CIFAR-10 classifier. 
Many popular DNN training optimizers, such as Stochastic Gradient Descent (SGD)~\cite{sgd}, SGD with Momentum~\cite{momentum} and Adam~\cite{adam}, utilize the learning rate in their optimization execution, indicating that the learning rate is a critical hyper-parameter for DNN training.
For example, for MNIST, TensorFlow chooses Adam optimizer with the fixed LR of 0.0001, and Caffe, Torch and Theano all choose SGD optimizer with a fixed LR, but they set their default choice for the fixed LR to 0.01, 0.05 and 0.1 respectively. In comparison, for CIFAR-10, TensorFlow, Torch and Theano choose SGD with fixed LR values of 0.1, 0.001 and 0.01 respectively while Caffe changes its LR function to a two-step decay LR policy with 0.001 as the LR value for the first 4,000 iterations in training and 0.0001 as the updated LR value for the last 1,000 iterations of training~\cite{GTDLBenchTSC}.
However, when a new DNN model is used for training or an existing DNN model is trained for a new learning task or a new dataset, domain-scientists and engineers have found it hard to select and compose a good learning rate policy and avoid worse LR choices for effective training of DNN models.
The manual-tuning task for finding a good or acceptable LR policy with respect to the training accuracy objective and training time constraint can be labor-intensive and error-prone, especially given the large search space of LR values for LR functions of either single-parameter or multi-parameters. There is a high demand for designing and developing a systematic approach to selecting and composing a good LR policy for a given learning task and dataset and a given DNN training algorithm. We argue that a good LR policy can notably improve the DNN training performance on both model test accuracy and model training time, significantly alleviate the frustration of manual-tuning difficulty and costs, and more importantly, can help avoiding the bad LR policy choices that will lead to below average or poor training performance.  

Bearing these objectives in mind, we present a systematic study of 15 representative learning rate functions from four LR algorithm families. This paper makes three original contributions. {\it First}, we develop an LR tuning mechanism to enable dynamically tuning and verification of LR policies with respect to desired accuracy goal and training time constraints, e.g., the pre-set \#Iterations or \#Epochs (the number of complete pass-throughs of the training data $\dataset$). {\it Second}, we develop an LR policy recommendation system (LRBench) to select and compose good LR policies from the same and/or different LR function(s) and avoid bad ones for a given learning task and a given dataset and DNN model. {\it Third}, we incorporate the support of different DNN optimizers and the recommendation of adaptive composite LR policy. The adaptive composite LR policy can further improve the quality of LR policy selection by enabling the creation of a multi-policy LR by combining multi-LR policies from different LR functions at different stages of the training process, boosting the overall performance of DNN model training on accuracy and training time.  
We evaluate our approach using four benchmark datasets, MNIST~\cite{mnistlenet}, CIFAR-10~\cite{cifar10-100}, SVHN~\cite{svhn} and ImageNet~\cite{ILSVRC}, and  three families of DNN backbone algorithms for model training: LeNet~\cite{mnistlenet}, CNN3~\cite{caffe}, and ResNet~\cite{resnet}. The results show that our approach is effective and the LR policies chosen by LRBench can consistently deliver high DNN model accuracy, outperform the existing recommended default LR policies for a given DNN model, learning task and dataset, and reduce the DNN training time by 1.6$\sim$6.7$\times$ to meet a targeted accuracy.

\section{Problem Statement} \label{problem-statement}
The DNN training with a given set of hyperparameters will output a trained model $F$ with dataset-specific model parameters ($\Theta$). During the training, an optimizer is used to update the model parameters and improve the model performance iteratively with two important optimizations. (1) A loss function ($L$) is computed statistically and used to measure the prediction deviation of the DNN model output to the ground truth, which enables the optimizer to reduce and minimize the loss value (error) throughout the iterative model update process.
(2) The learning rate policy ($\eta(t)$) is leveraged by the optimizer to control and adjust the amount of model parameter updates to be exercised during each training iteration $t$, which enables the optimizer to tune the rate of the update to the model parameters between slow and fast based on the specific learning rate value given at each training iteration. There are three primary goals for DNN training to adjust the extent of the update on the model parameters based on a specific LR policy: (i) to control the model learning speed, (ii) to avoid over-fitting to the single mini-batch, and (iii) to ensure that the model converges to global/local optimum.

Non-convex optimization algorithms are widely adopted as the optimizer for DNN training, such as Stochastic Gradient Descent (SGD)~\cite{sgd}, SGD with Momentum~\cite{momentum}, Adam~\cite{adam}, Nesterov~\cite{nesterov} and so forth.
For SGD, the DNN parameter update can be formalized as follows: 
\begin{equation}
\small
\Theta_{t+1} = \Theta_{t} - \eta(t) \nabla L
\label{formula:sgd}
\end{equation}
where $t$ represents the current iteration, $L$ is the loss function, $\nabla L$ is the gradients and $\eta(t)$ is the learning rate (LR) at iteration $t$ that controls the extent of the update to the model parameters (i.e., $\Theta_{t+1} - \Theta_{t} = -\eta(t) \nabla L$).

For other optimizers, such as SGD with Momentum (Momentum) and Adam, they adopt a similar method to update model parameters. For example, Momentum will update the model parameters $\Theta$ as Formula~(\ref{formula:momentum}) shows.
\begin{equation}
\small
V_{t} = \gamma V_{t-1} - \eta(t) \nabla L, \quad \Theta_{t+1} = \Theta_{t} + V_{t}
\label{formula:momentum}
\end{equation}
where $V_t$ is the accumulated gradients at iteration $t$ to be updated to the model parameters and $\gamma$ is a coefficient applied to the previous $V_{t-1}$, which is typically set to 0.9. Adam is another popular optimizer widely used in DNN training. It updates the model parameters $\Theta$ as Formula~(\ref{formula:adam}) shows.
\begin{equation}
\small
\begin{aligned}
M_t &= \beta_1 M_{t-1} + (1-\beta_1) \nabla L, \quad V_t = \beta_2 V_{t-1} + (1-\beta_2) (\nabla L)^2 \\
\hat{M_t} &= \frac{M_t}{1-\beta_1^t},  \quad \hat{V_t} = \frac{V_t}{1-\beta_2^t}, \quad \Theta_{t+1} = \Theta_{t} - \frac{\eta(t)}{\sqrt{\hat{V_i}}+\epsilon} \hat{M_t}
\end{aligned}
\label{formula:adam}
\end{equation}
where $\beta_1$ and $\beta_2$ are the coefficients to balance the previous accumulated gradients and the square of gradients. Typically, we will set $\beta_1=0.9$, $\beta_2=0.999$ and $\epsilon=10^{-8}$. Formula~(\ref{formula:sgd})$\sim$(\ref{formula:adam}) all contain the learning rate policy $\eta(t)$ as other popular optimizers do, such as Nesterov~\cite{nesterov} and AdaDelta~\cite{adadelta}. 

In addition, \cite{neural-optimizer-search} proposed to search the optimizers for training DNNs by modeling an optimizer as Formula~(\ref{formula:neural-optimizer-search})
\begin{equation}
\small
\Theta_{t+1} = \Theta_{t} - \eta(t) b(u_1(op_1), u_2(p_2))
\label{formula:neural-optimizer-search}
\end{equation}
where $op_1$ and $op_2$ are the operands, such as $\nabla L$, $M_t$ and $V_t$ in Formula~(\ref{formula:adam}), $u_1(.), u_2(.)$ and $b(.,.)$ denote the unary and binary functions respectively, such as mapping the input $x$ to $-x$ and $log|x|$ for the unary functions, and addition and multiplication for the binary functions. In particular, the learning rate policy $\eta(t)$ still plays a critical role in the searching and optimization process.

Learning rate optimization is a subproblem of hyper-parameter optimization that is only for learning rate $\eta$. For DNN training, given an optimizer $\mathcal{O}$ and a deep neural network $F_\Theta$ with trainable model parameters $\Theta$, the optimizer $\mathcal{O}$ minimizes the loss $L(x;F_\Theta)$ over i.i.d. samples $x$ from a natural (grand truth) distribution $\mathcal{G}_x$. In practice, the optimizer $\mathcal{O}$ will map a training dataset $\mathcal{X}^{train}$ to data-specific model parameters $\Theta$ for a given deep neural network $F_\Theta$, that is $\Theta = \mathcal{O}(\mathcal{X}^{train})$. An important hyper-parameter for the optimizer $\mathcal{O}$ is the learning rate policy $\eta=\eta(t)$. With the chosen $\eta$, we have the optimizer $\mathcal{O}_\eta$ and $\Theta = \mathcal{O}_\eta(\mathcal{X}^{train})$. Learning rate optimization aims at identifying a good learning rate policy $\eta$ to minimize the generalization error $\mathbb{E}_{x\sim\mathcal{G}_x}[L(x; F_{\mathcal{O}_\eta(\mathcal{X}^{train})})]$. In practice, we use a validation dataset $\mathcal{X}^{val}$ to estimate the generalization error, that is $L_{x\in\mathcal{X}^{val}}(x; F_{\mathcal{O}_\eta(\mathcal{X}^{train})})$.
$\mathcal{P}$ denotes the set containing all possible LR policies, the LR optimization problem is formalized as Formula~(\ref{formula:lr-optimization}):
\begin{equation}
\small
\hat{\eta} =  argmin_{\eta \in \mathcal{P}}L_{x\in\mathcal{X}^{val}}(x; F_{\mathcal{O}_\eta(\mathcal{X}^{train})})
\label{formula:lr-optimization}
\end{equation}

Different from other hyper-parameters, such as the weight decay rate, number of filters, kernel size, which are typically constant in the entire training process, the learning rate may change over the training iteration $t$. In practice, we choose a finite set of $S$ LR policies, consisted of different LR functions, denoted as $\mathcal{P}^*\subseteq\mathcal{P}$ and $\mathcal{P}^*=\{\eta^1(t), \eta^2(t), ..., \eta^S(t)\}$, e.g., $\eta^1(t)=k$ (a fixed LR) and $\eta^2(t)=\gamma^t$ ($\gamma<1$). Hence, we can formalize the LR optimization as Formula~(\ref{formula:lr-optimization-practice}). That is to select or compose the optimal LR policy from the candidate set $\mathcal{P}^*=\{\eta^1(t), \eta^2(t), ..., \eta^S(t)\}$.
\begin{equation}
\small
\hat{\eta} =  argmin_{\eta \in \mathcal{P}^*=\{\eta^1(t), ..., \eta^S(t)\}}L_{x\in\mathcal{X}^{val}}(x; F_{\mathcal{O}_\eta(\mathcal{X}^{train})})
\label{formula:lr-optimization-practice}
\end{equation}

\section{Learning Rate Selection And Composition}\label{section:lr-select-compose}
Learning rate is a function of the training iteration $t$ with a set of parameters and a method to determine the learning rate value at each iteration $t$ of the overall training process. A learning rate policy specifies a concrete parameter setting of an LR function. For example, a fixed learning rate of 0.01 is an LR policy of constant learning rate method with a fixed value of 0.01 throughout all iterations of the model training. Another example is the two-step LR policy of 0.01 in the first half of the training iterations and 0.001 in the second half of the training iterations.  In this section we will cover a total of 15 functions from three families of LR functions: fixed LRs, decaying LRs and cyclic LRs. We use the term of single learning rate policy to refer to the LR policy that corresponds to a single LR function, and refer to the LR policy that is defined by combining multiple LR policies from two or more LR functions as composite LR or multi-policy LR. We first describe our approach to select single LR policy for a given learning task, dataset and DNN backbone algorithm for model training. Then we introduce our composite LR scheme for selecting and composing an adaptive LR policy to further boost the overall training performance in terms of accuracy or training time given a target accuracy.

\subsection{Single Policy Learning Rates}
\noindent {\bf Fixed LRs (FIX)}, also called constant LRs, use a pre-selected fixed LR value throughout the entire training process, represented by $\eta(t) = k$ with $k$ as the only hyper-parameter to tune. However, too small $k$ value may slow down the training progress significantly. Too large $k$ value may accelerate the training progress at the cost of causing the loss function to fluctuate wildly, making the training fail to converge, resulting in very low accuracy. The conservative approach is popularly used, which uses a small value to ensure the model convergence and avoid oscillating loss (e.g., 0.01 for MNIST on LeNet and 0.001 for CIFAR-10 on CNN3). However, choosing a small and yet good fixed LR value is challenging. Even for the same learning task and dataset, e.g., CIFAR-10, different DNN models need to choose different constant values to meet the target accuracy goal (e.g., for CIFAR-10, 0.001 on CNN3 and 0.1 on ResNet-32). Another limitation of fixed LR policies is that it cannot adapt to the needs of different learning speeds during different training stages of the entire iterative learning process, and thus suffers from reaching the peak accuracy due to missing the speed-up opportunity when the training is on the plateau or failing to converge at the end of training.

\subsection{Composite Policy Learning Rates}
There are two types of composite learning rate schemes, according to whether the LRs are composed using the same LR function or using two or more different LR functions. The former is coined as the {\em homogeneous multi-policy} LRs and the later is called the {\em heterogeneous multi-policy} LRs. 

\noindent {\bf Decaying LRs\/} improve the limitation of the fixed LRs by using decreasing LR values during training. Similar to simulated annealing, training with a decaying LR starts with a relatively large LR value, which is reduced gradually throughout the training, aiming to accelerate the learning process while ensure that the training converges with good accuracy or meeting the target accuracy. A decaying LR policy is defined by a decay function $g(t)$ and a constant coefficient $k$, denoted by $\eta(t) = k g(t)$. $g(t)$ gradually decreases from the upper bound of 1 as the number of iterations ($t$) increases, and the constant $k$ value serves as the starting learning rate.

\begin{table}[h!]
\centering
\caption{Decaying Functions $g(t)$ for Decaying LRs}
\label{table:decaying-lr}
\small
\begin{tabular}{ccccc}
\hline
abbr. & $g(t)$ & Schedule & Param & \#Param \\ \hline
STEP & $\gamma^{floor(t/l)}$ & t, l & $\gamma, l$ & 2 \\ \hline
NSTEP & \small{\makecell{$l_0, ..., l_{n-1}$, \\ $\gamma ^i, \; i\in \mathbb{N}~s.t.$\\$l_{i-1} \le t < l_i$}} & $t, l_i$ & $\gamma,l_i$ & $n+1$ \\ \hline
EXP & $\gamma ^ t$ & $t$ & $\gamma$ & 1 \\ \hline
INV & $\frac{1}{(1+t\gamma)^p}$ & $t$ & $\gamma, p$ & 2 \\ \hline
POLY & \small{$(1-\frac{t}{max\_iter})^p$} & $t$ & $p$ & 1 \\ \hline
\end{tabular}
\end{table}

Table~\ref{table:decaying-lr} lists the 5 most popular decaying LRs, supported in LRBench. The STEP function defines the LR policy at iteration $t$ with 2 parameters, a fixed step size $l$ ($l>1$) and an exponential factor $\gamma$. The LR value is initialized with $k$ and decays every $l$ iterations by $\gamma$. The NSTEP enriches STEP by introducing $n$ variable step sizes, denoted by $l_0, l_1, ..., l_{n-1}$, instead of the one fixed step size $l$. NSTEP is initialized by $k$ ($g(t)=1$ when $i=0, t<l_0$) and computed by $\gamma ^i$ (when $i>0$ and $l_{i-1} \le t < l_i$). EXP is an LR function defined by an exponential function ($\gamma ^ t$).
Although EXP, STEP and NSTEP all use an exponential function to define $g(t)$, their choice of concrete $\gamma$ is different. 
To avoid the learning rate decaying too fast due to the exponential explosion, EXP uses a $\gamma$ that is close to 1, e.g., 0.99994 and reduces the LR value every iteration. In contrast, STEP and NSTEP employ a small $\gamma$, e.g., 0.1, and decay the LR value using one fixed step size $l$ or using $n$ variable step sizes $l_i$. The total number of steps is determined, for STEP, by the step size and the pre-defined training \#Iterations (or \#Epochs), and for NSTEP, $n$ is typically small, e.g., 2$\sim$5 steps. Other decaying LRs are based on the inverse time function (INV) and the polynomial function (POLY) with parameter $p$ as shown in Table~\ref{table:decaying-lr}. A good selection of the value settings for these parameters is challenging and yet critical for achieving effective training performance when using decaying LR policies.

Several studies~\cite{cifar10-100,Bengio-PracticalRecommendations,resnet} show that with a good selection of the decaying LR policies, they are more effective than the fixed LRs for improving training performance on accuracy and training time. However, \cite{clr,superconvergence,sgdr,LRBenchBigData} recently show that decaying LRs tend to miss the opportunity to accelerate the training progress on the plateau in the middle of training when initialized with too small LR values and result in slow convergence and/or low accuracy. While larger initial values for decaying LRs can lead to the same problem as those for fixed LRs.

STEP is an example of homogeneous multi-policy LRs created using a single LR function FIX by employing multiple fixed LRs defined by the LR update schedule based on training iteration $t$ and step size $l$. NSTEP is another example of homogeneous multi-policy LRs by employing $n$ different FIX policies, each changing the LR value based on $t$ and the $n$ different step size: $l_i$ ($i=0,\dots, n-1$).

\noindent {\bf Cyclic LRs (CLRs)} are proposed recently by~\cite{clr,superconvergence,sgdr}, to address the above issue of decaying LRs. CLRs by design change the LR cyclically within a pre-defined value range, instead of using a fixed value or reducing the LR by following a decaying policy, and some target accuracy thresholds can be achieved earlier with CLRs in shorter training time and smaller \#Epochs or \#Iterations. In general, cyclic LRs can be defined by $\eta(t) = |k_0 - k_1| g(t) + min(k_0, k_1)$, where $k_0$ and $k_1$ specify the upper and lower value boundaries, $g(t)$ represents the cyclic function whose domain ranges from 0 to 1, and $min(k_0, k_1) \le \eta(t) \le max(k_0, k_1)$.
For each CLR policy, three important parameters should be specified: $k_0$ and $k_1$, which specify the initial cyclic boundary, and the half-cycle length $l$, defined by the half of the cycle interval, similar to the step size $l$ used in decaying LRs. The good selection of these LR parameter settings is challenging and yet critical for the DNN training effectiveness under a cyclic LR policy (see Section~\ref{section:experimental-analysis} for details).

\begin{table}[h!]
\centering
\caption{Cyclic Functions $g(t)$ for Cyclic LRs}
\label{table:cyclic-lr}
\small
\begin{tabular}{ccccc}
\hline
abbr. & $g(t)$ & Schedule & Param & \#Param \\ \hline
TRI & \small{\makecell{$TRI(t)=\frac{2}{\pi}|arcsin(sin(\frac{\pi t}{2l}))|$}} & $t, l$ & $l$ & 1 \\ \hline
TRI2 & \small{$\frac{1}{2^{floor(\frac{t}{2l})}}TRI(t)$} & $t, l$ & $l$ & 1 \\ \hline
TRIEXP & $\gamma^{t}TRI(t)$ & $t, l$ & $\gamma, l$ & 2 \\ \hline
COS & \makecell{$COS(t)=\frac{1}{2}(1+cos(\pi \frac{t}{l}))$} & $t, l$ & $l$ & 1\\ \hline
SIN & \small{\makecell{$SIN(t)=|sin(\pi\frac{t}{2l})|$}} & $t, l$ & $l$ & 1 \\ \hline
\end{tabular}
\end{table}

We support three types of CLRs currently in LRBench: triangle-LRs, sine-LRs and cosine-LRs as Table \ref{table:cyclic-lr} shows. 
TRI is formulated with a triangle wave function $TRI(t)$ bounded by $k_0$ and $k_1$. TRI2 and TRIEXP are two variants of TRI by multiplying $TRI(t)$ with a decaying function, $\frac{1}{2^{floor(\frac{t}{2l})}}$ for TRI2 and $\gamma^{t}$ for TRIEXP. TRI2 reduces the LR boundary ($|k_0 - k_1|$) every $2 l$ iterations while TRIEXP decreases the LR boundary exponentially.~\cite{sgdr} proposed a cosine function with warm restart and can be seen as another type of CLRs, and we denote it as COS. We implement COS2 and COSEXP as the two variants of COS corresponding to TRI2 and TRIEXP in LRBench and also implement SIN, SIN2 and SINEXP as the sine-CLRs in LRBench. 

\begin{figure}[h!]
\centering
\subfloat[At the 70th iteration]{
  \centering
  \includegraphics[trim=60 25 20 25, clip,width=0.25\textwidth]{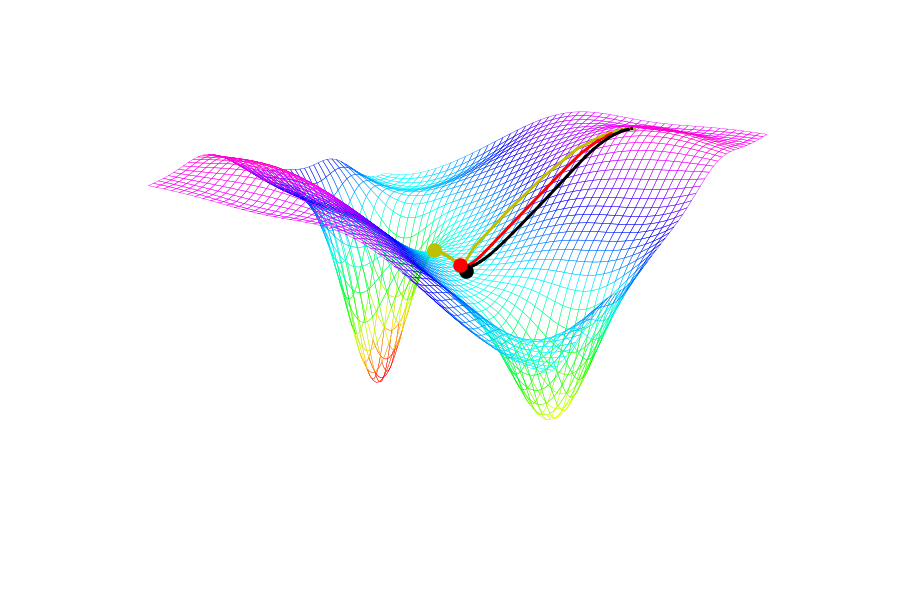}
  \label{fig:fix-nstep-triexp-70}
} 
\hspace{-4mm}
\subfloat[At the 100th iteration]{
  \centering
  \includegraphics[trim=60 25 20 25, clip,width=0.25\textwidth]{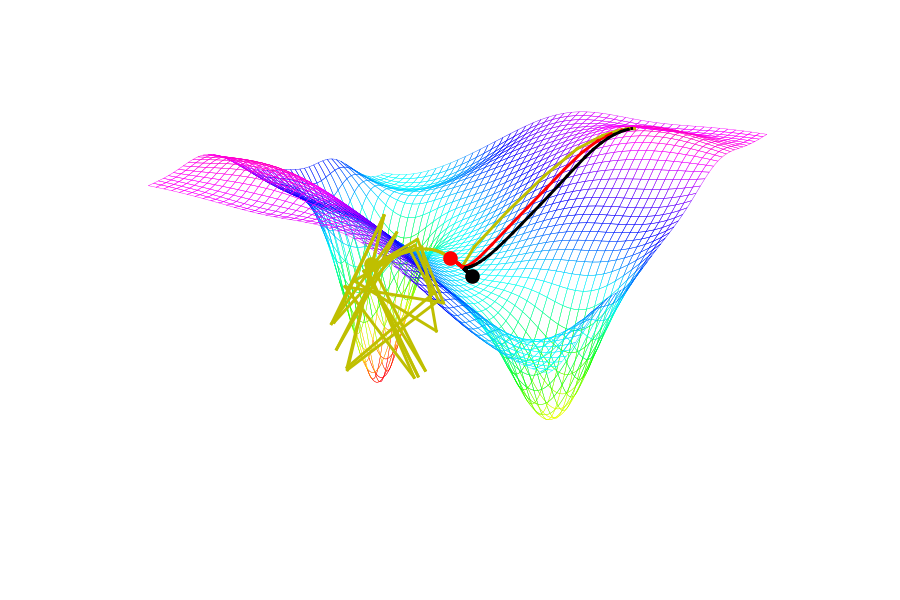}
  \label{fig:fix-nstep-triexp-100}
} 
\hspace{-4mm}
\subfloat[At the 120th iteration]{
  \centering
  \includegraphics[trim=60 25 20 25, clip,width=0.25\textwidth]{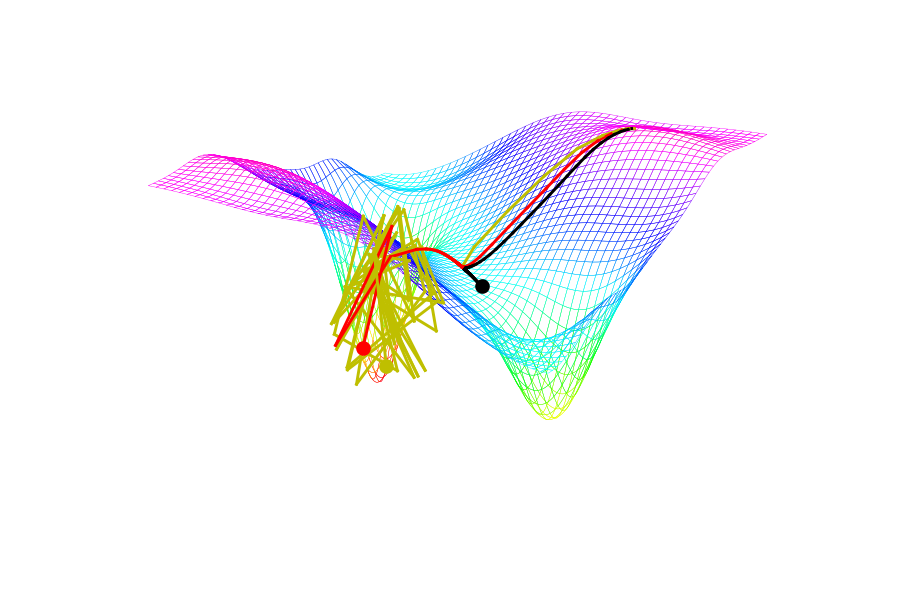}
  \label{fig:fix-nstep-triexp-120}
} 
\hspace{-4mm}
\subfloat[At the 149th iteration]{
  \centering
  \includegraphics[trim=60 25 20 25, clip,width=0.25\textwidth]{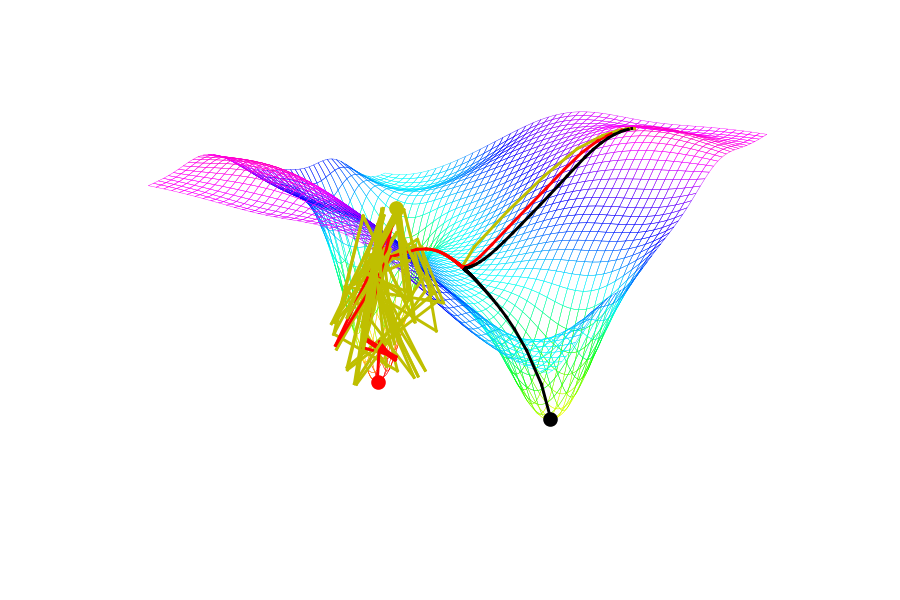}
  \label{fig:fix-nstep-triexp-149}
} 
\caption{\small{Visualization of the Training Process with Different LRs}}
\label{fig:visulization-fix-nstep-triexp}
\end{figure}

We visualize the optimization process of three LR functions, FIX, NSTEP and TRIEXP, from the above three categories in Figure~\ref{fig:visulization-fix-nstep-triexp}. The corresponding LR policies are FIX($k=0.025$) in black, NSTEP($k=0.05, \gamma=0.5, l=[120, 130]$) in red and TRIEXP($k_0=0.05, k_1=0.25, \gamma=0.9, l=25$) in yellow. All three LRs start from the same initial point, and the optimization process lasts for 150 iterations. The color on the grid marks the value of the cost function to be minimized, where the global optimum corresponds to red. In general, we observe that different LRs lead to different optimization paths. Although three LRs exhibit little difference in terms of the optimization path up to the 70th iteration, the FIX lands at a different local optimum at the 149th iteration rather than the global optimum (red). It shows that the accumulated impacts of LR values could result in sub-optimal training results. Also, at the beginning of the optimization process, TRIEXP achieved the fastest progress according to Figure~\ref{fig:fix-nstep-triexp-70} and~\ref{fig:fix-nstep-triexp-100}. This observation indicates that starting with a high LR value may help accelerate the training process. However, the relatively high LR values of TRIEXP in the late stage of training also introduce high ``kinetic energy'' into this optimization and therefore, the cost values are bouncing around chaotically as the TRIEXP optimization path (yellow) shows in Figure~\ref{fig:fix-nstep-triexp-100}$\sim$\ref{fig:fix-nstep-triexp-149}, where the model may not converge. This example also illustrates that simply using a single LR function may not achieve the optimal optimization progress due to the lack of flexibility to adapt to different training states.

Studies by~\cite{clr,superconvergence,sgdr,LRBenchBigData} confirm independently that increasing the learning rate cyclically during the training is more effective than simply decaying LRs continuously as training progresses. On the one hand, with relatively large LR values, cyclic LRs can accelerate the training progress when the model is trapped on the plateau and escape the local optimum by updating the model parameters more aggressively. On the other hand, cyclic LRs, especially decaying cyclic LRs (i.e., TRI2, TRIEXP) can further reduce the boundaries of LR values, as training progresses through different stages, such as the training initialization phase, the middle stage of training in which the model being trained is trapped on a plateau or a local optimum, and the final convergence phase. We argue that the learning rate should be actively changed to accommodate to different training phases. However, existing learning rates, such as the fixed LRs, decaying LRs and cyclic LRs are all defined by one function with one set of parameters. Such design limits its adaptability to different training phases and often results in slow training and sub-optimal training results.

According to our characterization of single-policy and multi-policy LRs, we can view TRI2, COS2 and SIN2 as examples of homogeneous multi-policy LRs, because they are created by combining multiple LR policies through a simple and straightforward integration of decaying LR policy to refine the cyclic LR policy using decaying ($k_0, k_1$) by one half every $2 l$ iterations. We have analyzed how and why these multi-policy LRs provide more flexibility and adaptability in selecting and composing a multi-policy LR mechanism for more effective training of DNNs in this section. Our empirical results also confirm consistently that multi-policy LRs hold the potential to further improve the test accuracy of the trained DNN models.

\begin{table}[h!]
\centering
\caption{Different LR Policies for CIFAR-10 on ResNet-32, showing the benefit of advanced composite LRs.}
\label{table:lr-comparison-cifar10-resnet32}
\small
\scalebox{0.9}{
\begin{tabular}{c|c|cc}
\hline
\multicolumn{2}{c|}{LR Policy}                                                         & \multirow{2}{*}{\#Iter} & \multirow{2}{*}{Accuracy (\%)}  \\ \cline{1-2}
Category                             & LR Function                                  &                             &                                            \\ \hline
Fixed LR                       & FIX ($k$=0.1)      & 61000                                               & 86.08                                              \\ \hline
Decaying LR             & NSTEP ($k$=0.1,$\gamma$=0.1,$l$={[}32000,48000{]})        & 53000            & 92.38$\pm$0.04                                        \\ \hline
Cyclic LR                  & SINEXP ($k_0$=0.0001,$k_1$=0.9,$\gamma$=0.99994)           & 64000              & 92.81$\pm$0.08                                        \\ \hline
\multirow{3}{*}{\makecell{Advanced\\Composite LR}} & 0-30000 iter:TRI ($k_0$=0.1, $k_1$=0.5, $l$=1500)           & \multirow{3}{*}{64000}& \multirow{3}{*}{\textbf{92.91}}            \\ \cline{2-2}
                                  & 30000-60000 iter: TRI ($k_0$=0.01, $k_1$=0.05, $l$=1000)                           &   &                                                     \\ \cline{2-2}
                                  & 60000-64000 iter: TRI ($k_0$=0.001, $k_1$=0.005, $l$=500)                   &      &                                                        \\ \hline
\end{tabular}
} 
\end{table}

\begin{figure*}[h!]
\centering
\subfloat[Decaying LR (NSTEP)]{
  \centering
  \includegraphics[width=0.3\textwidth]{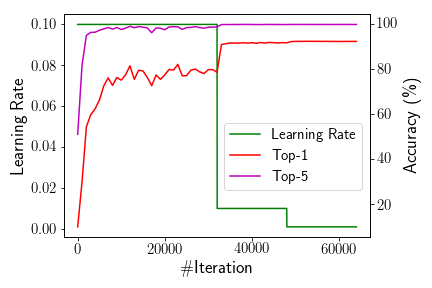}
  \label{fig:nstep-cifar10-resnet32}
} 
\subfloat[Cyclic LR (SINEXP)]{
  \centering
  \includegraphics[width=0.3\textwidth]{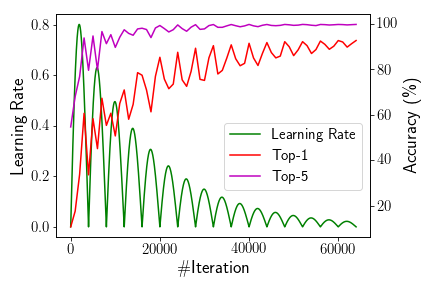}
  \label{fig:sinexp-cifar10-resnet32}
} 
\subfloat[Multi-policy LR]{
  \centering
  \includegraphics[width=0.3\textwidth]{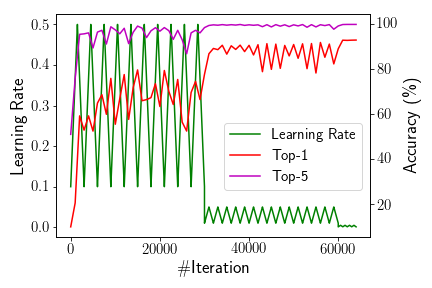}
  \label{fig:hlr-cifar10-resnet32}
} 
\caption{Comparison of NSTEP, SINEXP and Heterogeneous Multi-policy LRs (CIFAR-10, ResNet-32), showing the benefit of advanced composite LR in performance improvement during the latter half of the training.}
\label{fig:lr-comparison-cifar10-resnet32}
\end{figure*}

\noindent {\bf Advanced Composite LRs.\/} There are different combinations of multi-policy LRs that can be beneficial for further boosting the DNN model training performance. Consider a new multi-policy LR mechanism, it is created by composing three triangle LRs in three different training stages with decaying cyclic upper and lower bounds and varying decaying step sizes over the total training iterations. Table~\ref{table:lr-comparison-cifar10-resnet32} shows the effectiveness of this multi-policy LR policy through a comparative experiment on ResNet-32 with CIFAR-10 dataset using Caffe DNN framework. The experiment compares four scenarios. (1) The single-policy LR with the fixed value of 0.1 achieves accuracy of 86.08\% in 61,000 iterations out of the default total training iterations of 64,000. (2) The decaying LR policy of NSTEP function with specific $k, \gamma, l$ achieves the test accuracy of 92.38\% for trained model with only 53,000 iterations out of 64,000 total iterations. (3) The cyclic LR policy of SINEXP function with specific LR parameters ($k_0, k_1, \gamma$) achieves the accuracy of 92.81\% at the end of 64,000 total training iterations. (4) The advanced composite multi-policy LR mechanism is created by LRBench for CIFAR-10 (ResNet-32), which achieves the highest test accuracy of 92.91\% at the total training rounds of 64,000, compared to other three LR mechanisms. 
Figure~\ref{fig:lr-comparison-cifar10-resnet32} further illustrates the empirical comparison results. It shows the learning rate (green curves) and Top-1/Top-5 accuracy (red/purple curves) for the top three performing LR policies: NSTEP, SINEXP and the advanced composite LR.

We highlight three interesting observations.
{\it First,} this multi-policy LR achieved the highest accuracy of 92.91\%, followed by the cyclic LR (SINEXP, 92.81\%) and decaying LR (NSTEP, 92.38\%) while the fixed LR achieved the lowest 86.08\% accuracy.
{\it Second,} from the LR value y-axis (left) and the accuracy y-axis (right) in Figure~\ref{fig:nstep-cifar10-resnet32} and Figure~\ref{fig:sinexp-cifar10-resnet32}, we observe that the accuracy increases as the overall learning rates decrease over the iterative training process ($t$ on x-axis). However, without dynamically adjusting the learning rate values in a cyclic manner, NSTEP missed the opportunity to achieve higher accuracy by simply decaying its initial LR value over the iterative training process.
{\it Third,} Figure~\ref{fig:sinexp-cifar10-resnet32} shows that even with a cyclic LR policy such as SINEXP, it fails to compete with the new multi-policy LR that we have designed because it fails to decay faster and at the end of the training iterations, it still uses much higher learning rates. In comparison, Figure~\ref{fig:hlr-cifar10-resnet32} shows the training efficiency using our new multi-policy LR for CIFAR-10 on ResNet-32. It uses three different CLR functions in three different stages of the overall 64,000 training iterations, each stage reducing the cyclic range by one-tenth decaying of $k_0$ and $k_1$ and reducing the step size $l$ by 500 iterations.

Recent research has shown that variable learning rates are advantageous by empowering training with adaptability to dynamically change the LR value throughout the entire training process to adapt to the need of different learning modes (learning speed, step size and direction for changing learning speed) at different training phases. A key challenge for defining a good multi-policy LR, be it homogeneous or heterogeneous, is related to the problem of how to determine the dynamic settings of its parameters, such as $l$ and $k_0, k_1$. Existing hyper-parameter search tools are typically designed for tuning the parameters at the initialization of the DNN training, which will not change during training, such as the number of DNN layers, the batch size, the number of iterations/epochs, the optimizer (SGD, Adam, etc.). Thus, an open problem of selecting and composing LR policies with evolving LR parameters is to develop a systematic approach to selecting and composing LRs. In the next section, we present our framework, called LRBench, to create a good LR policy by dynamically tuning, composing and selecting a good LR policy for a learning task and a given dataset on a chosen DNN model. 

\section{Design Overview}
\begin{figure*}[h!]
    \centering
    \includegraphics[width=0.7\textwidth]{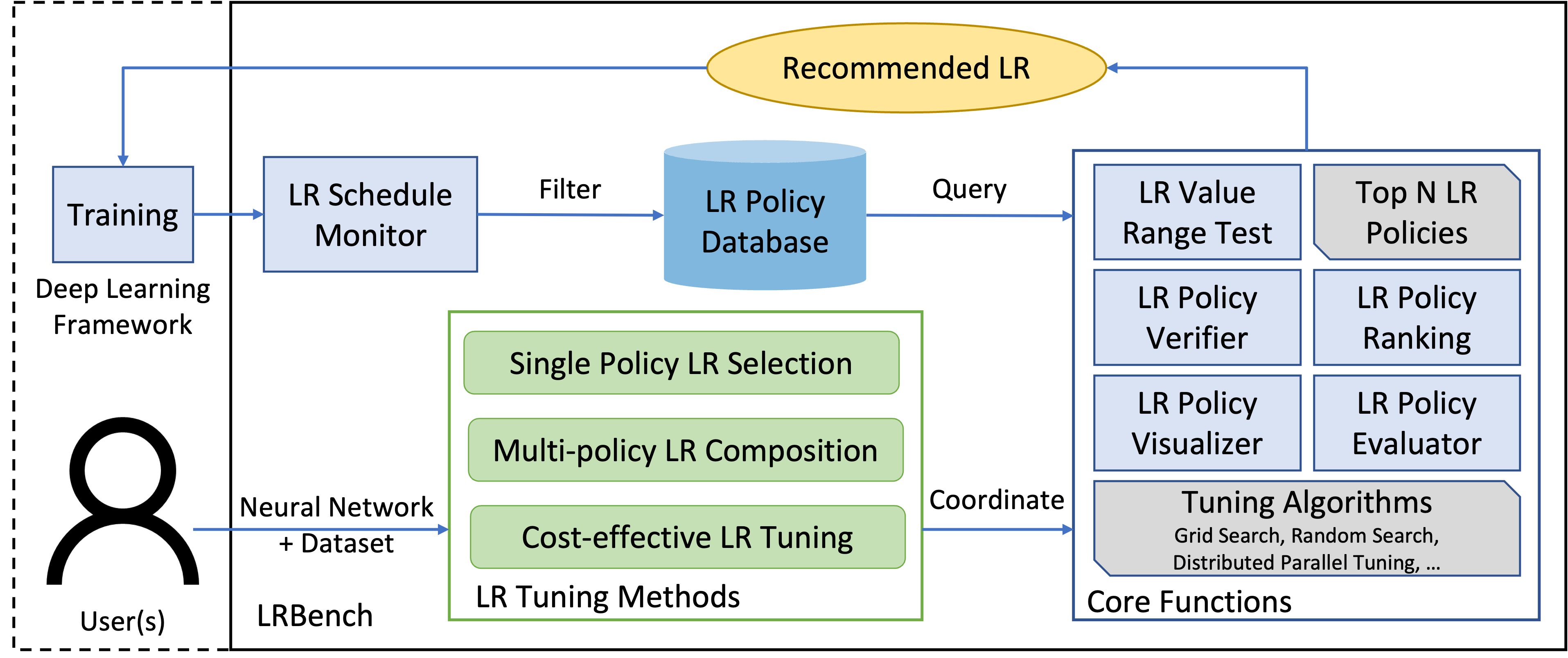}
\caption{LRBench Architecture Overview}
\label{fig:lrbench-overview}
\end{figure*}

LRBench is a \textbf{L}earning \textbf{R}ate \textbf{Bench}marking system for tuning, composing and selecting LR policies for DNN training. The current prototype of LRBench implemented 15 LR functions in 4 families of LR policies and 8 evaluation metrics, covering utility, cost and robustness of LR policies defined by~\cite{LRBenchBigData}. Figure~\ref{fig:lrbench-overview} shows the architecture of LRBench, consisting of 8 functional components. (1)~The LR schedule monitor tracks the training status and triggers the LR value update based on the specific LR policy. (2) The LR policy database stores the empirical LR tuning results organized by the specific neural network and dataset. (3) The LR value range test will estimate a good LR value range to reduce the LR search space. (4) The LR policy verifier will perform the verification of the given LR policy in terms of whether it can meet the desired target accuracy or achieve optimal training performance. (5) The LR policy visualizer will help users visualize the LR values and DNN training progress. (6) The LR policy ranking algorithm will select the top $N$ ranked LR policies based on the empirical LR tuning results from the LR policy database for a pre-defined utility, cost or robustness metric. (7) The LR policy evaluator estimates the good LR parameters, such as the step size $l$, for evaluating and comparing alternative LR policies and dynamically tuning LR values. (8) The tuning algorithms in LRBench are provided by Ray Tune~\cite{ray-tune}, including grid search, random search and distributed parallel tuning algorithms to effectively explore the LR search space.

In practice, users can leverage a set of functional APIs provided by LRBench in their DNN training. For example, LRBench provides three main LR tuning methods: (1) single policy LR selection, (2) multi-policy LR composition and (3) cost-effective LR tuning. The first two LR tuning methods aim at achieving the highest possible model accuracy under a training time constraint, such as the number of training iterations. The third tuning method targets at reducing the training time cost for achieving a desired target accuracy. In some cases, such as the pruning in neural network search~\cite{neuralnetworksearch}, the goals for DNN training is to achieve a certain accuracy threshold, where a low training cost, such as using a small number of iterations, will significantly improve the training efficiency. 
LRBench leverages two complementary principles in LR tuning that is \textbf{exploration} and \textbf{exploitation}. Exploration helps LRBench explore different LR functions and LR parameters in DNN training, which can avoid falling into sub-optimal LRs or even bad ones. Exploitation allows LRBench to follow a few good LR policies and verify whether these LRs can help the DNN converge to high accuracy under the given training time constraints. A proper balance between these two principles is crucial for efficient LR tuning. For example, the single policy LR selection restricts the LR to a single policy and puts more effort into exploitation, compared to the multi-policy LR composition, which allows composing multiple LR policies and emphasizes more on exploration.

\noindent {\bf Implementation Details.} We have open-sourced LRBench on GitHub at \url{https://github.com/git-disl/LRBench}, which is written in Python and has a flexible modular design that allows users to leverage different modules to perform learning rate tuning. The frontend and LR policy database are implemented with Django and PostgreSQL respectively. LRBench supports popular deep learning frameworks, such as Caffe, PyTorch, Keras and TensorFlow through their Python APIs.

In this section, we describe LR tuning and LR verification, which are the two important features in LRBench to support automatic creation of new multi-policy LRs in the scenarios where the target accuracy is not met or can be improved by LRBench during its LR verification process.

\subsection{Learning Rate Tuning}\label{section:lr-tuning}
The learning rate is an important hyper-parameter, critical for efficient DNN training. Even for the fixed LR, it is non-trivial to find a good value. Advanced LRs, such as cyclic LRs and composite LRs introduce more learning rate parameters to control the LR values, making this process even harder. LRBench is designed to mitigate these issues with three important components, the LR Policy Database, LR Value Range Test and LR Schedule Monitor.

\noindent {\bf LR Policy Database} stores the empirical LR policy tuning results, including good LR value ranges, organized by the specific dataset, the specific learning task, the chosen DNN model (e.g., LeNet, ResNet, AlexNet, etc.), and the deep learning framework.
With the LR policy database, LRBench is able to rank existing LR policies based on a set of metrics, such as the loss, accuracy and training cost, and recommend a good LR policy for DNN training for a learning task on a given dataset with a chosen DNN model. It does so by first searching the LR policy database, and if the relevant LR policies are found, LRBench will use the stored LR policies, denoted as $\hat{\mathcal{P}} \subseteq \mathcal{P}$, as the starting point for performing LR tuning. Upon identifying a set of three good LR policies that can meet the target accuracy within the pre-defined training iterations, LRBench will select the highest ranked LR policy as the tuning output. Otherwise, LRBench will trigger the new LR policy generation process, which will perform the LR value range test to find the range of good LR values.

\begin{wrapfigure}{r}{0.45\linewidth}
    \centering
    \includegraphics[width=0.45\textwidth]{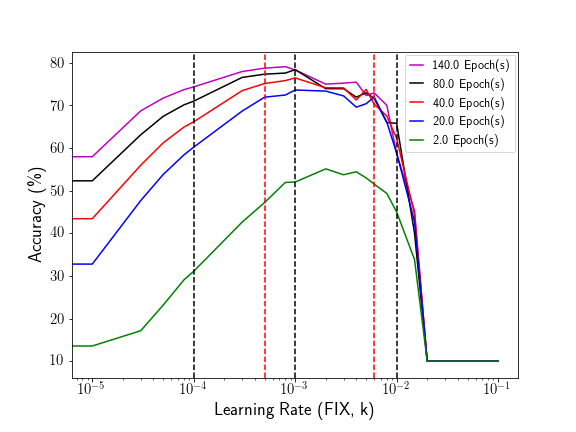}
\caption{Acc with Varying $k$ (FIX, CNN3, CIFAR-10)}
\label{fig:acc-lr-cifar10-fix}
\end{wrapfigure}

\noindent {\bf LR Value Range Test.\/}
The goal of LR value range test is to significantly reduce the search space of LR parameters for finding the good candidate LR policies. We first illustrate by example how to determine the proper ranges for training CNN3 on CIFAR-10.
Figure~\ref{fig:acc-lr-cifar10-fix} shows the results of Top-1 accuracy (y-axis) by varying the fixed LR values in a base-10 log scale (x-axis). The two red vertical dashed lines represent two fixed LR values: $k=0.0005$ and $0.006$, and the three black dashed lines represent another three fixed LR values: $k=0.0001$, $0.001$ and $0.01$  The different colors of curves correspond to the Top-1 accuracy obtained when training using different total \#Epochs. We observe from Figure \ref{fig:acc-lr-cifar10-fix} that the appropriate LR ranges is $[0.0005, 0.006]$ marked by the two red dashed lines, and the accuracy drops much faster after $k=0.006$. Thus, setting the upper bound of the LR range to be around 0.006 is recommended. Similarly, the lower bound by $k=0.0005$ is the LR value in which the accuracy increase is relatively stabilized, indicating that $[0.0005, 0.006]$ is a good LR value range for CNN3 training on CIFAR-10.
Another interesting observation is that with five different accuracy curves obtained using five different training \#Epochs (in five colors), the general accuracy
trend over varying LR values (x-axis) reveals similar patterns. This shows that it is sufficient for LRBench to examine the LR value range using a small number of variations in \#Epochs instead of enumerating a large number of possible \#Epochs, enabling LRBench to accelerate this process, which can also be automated by using grid or random search algorithms in LRBench. In comparison, the total of 140 epochs is used in Caffe as the recommended training time constraint for training CNN3 on CIFAR-10 (the default). This verification and tuning process also enables LRBench to remove worse LR policies through a small number of trial-and-error operations. For instance, by investigating the relationship between test accuracy and the good LR range, we can choose the fixed LR value from a small set of good LR values and significantly reduce the search space of finding a good LR policy by 99.4\% for CIFAR-10, computed by $(1-(0.006-0.00005))\times100\%/1$, compared to using the search space defined by the domain $[0, 1]$ of any given LR function for finding a good LR value under a given training \#Epochs.

\begin{algorithm}
\caption{Change-LR-On-Plateau}
\label{alg:changelronplateau}
\footnotesize
\begin{algorithmic}
    \Procedure{Change-LR-On-Plateau}{$m, \mathcal{P}, i$, training stop condition}
    \State \textbf{Input}: (1) the monitored metrics $m$, such as the training loss; (2) the chosen $n$ LR functions $\mathcal{P}=\{p^1(t), p^2(t), ..., p^n(t)\}~s.t.~p^1(t)\ge p^2(t) \ge ... \ge p^n(t)$; (3) the starting LR function index $i$; (4) the training stop condition, such as a pre-defined \#Iterations.
    \State \textbf{Output}: The trained DNN model $F_\Theta$.
    \LineComment{$\mathcal{M}$: store monitored metrics, $t$: the current iteration}
    \State Initialize  $t=0$, $\eta=p^i(t)$, $\mathcal{M}=\{\}$;
    \State Initialize the DNN model $F_\Theta$;
    \While{the training stop condition is not met}
        \State training $F_\Theta$ with $\mathcal{O}_\eta(\mathcal{X}^{train})$ for one iteration;
        \State obtain the monitored metrics $m$;
        \State action = Is-Trapped-On-Plateau-Action($\mathcal{M}, m, t)$;
        \If{action == ``increase''}
        \State $i=i-1$;
        \EndIf
        \If{action == ``decrease''}
        \State $i=i+1$;
        \EndIf
        \State $\mathcal{M}$.append($m$); $t = t + 1$;  
    \EndWhile\\
    \Return $F_\Theta$;
    \EndProcedure
\end{algorithmic}
\end{algorithm}

\noindent {\bf LR Schedule Monitor} is a system facility that is used for both LR tuning and LR verification. It monitors the training status, such as the training/validation loss, and dynamically changes the LR value to ensure effective training for meeting the target accuracy. Reduce-LR-On-Plateau is a popular method to tune LR on-the-fly during training included in~\cite{reducelronplateau}, which monitors the training/validation loss during the DNN training, when the learning stagnates on plateau, e.g., no improvement for the loss, Reduce-LR-On-Plateau will reduce the LR value to facilitate DNN training. However, Reduce-LR-On-Plateau is limited to multiple decreasing fixed LR values and the FIX LR function. When the training is trapped on a plateau, increasing LR by a cyclic LR function or even using a decaying LR function can be more beneficial and lead to faster training convergence and higher accuracy according to empirical studies~\cite{superconvergence,saddlepointissue}. In LRBench, we implement the Change-LR-On-Plateau method as Algorithm~\ref{alg:changelronplateau} shows to compose multi-policy LRs. Our Change-LR-On-Plateau method allows changing the current LR value at different phases of the training based on the specific LR function ($p(t)$) used, be it a decaying LR function or a cyclic LR function, rather than being limited to only the decreasing fixed LR values as Reduce-LR-On-Plateau. $i$ indicates the current LR function index in the input LR functions $\mathcal{P}$ (ordered in the descending order) for the learning rate value $\eta$, which will be used in the optimizer $\mathcal{O}_\eta(\mathcal{X}^{train})$. $\mathcal{M}$ will keep track of the monitored metrics $m$, and $t$ indicates the current training iteration. Is-Trapped-On-Plateau-Action($\mathcal{M}, m, t)$ will take these three variables and judge whether the current training is trapped on the plateau, such as for a certain number of consecutive training iterations, e.g., 5 iterations, whether the improvement of the monitored metrics are beyond a threshold, e.g., 0.05 for the training loss. Then, if the current training is not trapped on plateau, it will just return no action. Otherwise, it will return specific actions to change the index $i$ to switch between the LR functions. For example, during the beginning and middle of the training (i.e., $t$ is smaller than a pre-defined threshold), Is-Trapped-On-Plateau-Action($\mathcal{M}, m, t)$ will return ``increase'' to increase the LR value with $i=i-1$, allowing the training to quickly escape the plateau or local optimum, and when the training approaches the end of the pre-defined training iterations, it will return ``decrease'' with $i=i+1$ to reduce the LR value to help converge the DNN training.

\subsection{Learning Rate Verification}
When a new DNN model is chosen for an existing dataset and learning task such as CIFAR-10, even we use the same DL framework such as Caffe or TensorFlow, the system recommended LR policy for CIFAR-10 on AlexNet is no longer a good LR policy for ResNet. Similarly, the LR policy for MNIST with the 10-class classifier is not a good policy for CIFAR-10 either with only 77.26\% accuracy compared to 81.61\% of the Caffe default NSTEP policy on CNN3. This motivates us to incorporate a new functional component, called the LR policy verification, in our LRBench.

Our LR policy verification is carried out in three validation phases. In the first validation phase, we utilize the DNN training to perform the verification of the given LR policy in terms of whether it meets the desired target accuracy given by the user. If yes, then this policy is verified. Otherwise, we enter the second validation phase, which searches the LR policy database according to the learning task, the dataset, or the DNN model and select the top three ranked LR policies based on the Top-1 accuracy, for example. Other ranking metrics, such as Top-5 accuracy, training cost and robustness of the trained model, can also be employed in conjunction with Top-1 accuracy. Upon identifying the top three good LR policies, we select the highest ranked LR policy and compare it with the LR policy being verified. By good, we mean that all meet the target accuracy within the pre-defined training time constraints (e.g., \#Iterations or \#Epochs). If the LR policy being verified is no longer the winner in ranking, the verification will output the top ranked LR policy as the LRBench recommended LR policy as the replacement. The verification process is terminated. If the policy being verified remains to be the winner, the verification enters the last phase, in which LRBench will trigger the new LR policy generation process, which will perform the LR value range test. It aims to effectively reduce the search space of LR parameters for finding the good candidate LR policies through several mechanisms, such as the grid search or random search algorithms in LRBench.
A challenge for the third verification phase is to estimate the good LR value based on the current or the recent past model parameters. One approach is to use the optimal learning rate (denoted as M-opt LR) of the three most recent intermediate learning outcomes across two steps. Let $\Theta_t$ denote the model parameter in the current training iteration $t$, $\eta_{t+1}$ be the actual LR value used for the iteration $t+1$, and $\eta^*$ denote the calculated LR value across two consecutive model parameter updates, i.e., $(\Theta_{t+1}-\Theta_{t})$ and $(\Theta_{t+2}-\Theta_{t+1})$. One can estimate the M-opt $\eta^*$ using the following formula~\cite{superconvergence} on $\Theta_t$, $\Theta_{t+1}$ and $\Theta_{t+2}$:
\begin{equation}
\small
\begin{aligned}
\eta^* &= \eta_{t+1} \frac{||\Theta_{t+1} - \Theta_{t}||_1}{||(\Theta_{t+1}-\Theta_t)-(\Theta_{t+2}-\Theta_{t+1})||_1}\\
&= \eta_{t+1} \frac{||\Theta_{t+1} - \Theta_{t}||_1}{||2\Theta_{t+1}-\Theta_t-\Theta_{t+2}||_1}
\end{aligned}
\label{formula:m-opt-lr}
\end{equation}

The intuition of Formula~(\ref{formula:m-opt-lr}) is from the Newton's method for optimization, which will use the Hessian matrix (second derivative) $H_{L}$ of the loss function $L$ to update the model parameters as Formula~(\ref{formula:newton-optimization}) shows. The high computation cost of the Hessian matrix makes it infeasible to obtain the exact Hessian matrix for real optimization. \cite{hassian-free} proposed the Hessian-free optimization to obtain an estimate of the Hessian $h_{L}(\vartheta) \in H_{L}$ where $\vartheta$ is a single variable in $\Theta$ from two gradients as Formula~(\ref{formula:hessian-matrix-estimate}) shows and $\delta$ should be in the direction of the steepest descent. Based on the estimation, \cite{adasecant} further proposed an efficient learning rate calculated with Formula~(\ref{formula:adasecant-lr}) on the single variable $\vartheta \in \Theta$. From Formula~(\ref{formula:sgd}) on SGD, we have Formula~(\ref{formula:step-1-1}) and (\ref{formula:step-1-2}) on the single variable $\vartheta$. With Formula~(\ref{formula:adasecant-lr}), (\ref{formula:step-1-1}) and~(\ref{formula:step-1-2}), we derive Formula~(\ref{formula:step-2}) based on $\eta_{t+1} \approx \eta_{t}$, indicating the effective learning rate for a single variable $\vartheta \in \Theta$. Hence, for estimating the optimal learning rate for $\Theta$, we applied the L-1 normalization in Formula~(\ref{formula:m-opt-lr}) as~\cite{superconvergence} suggests.

\begin{equation}
\small
\Theta_{t+1} = \Theta_{t} -  H_{L}^{-1} \nabla L
\label{formula:newton-optimization}
\end{equation}

\begin{equation}
\small
h_{L}(\vartheta) = \lim_{\delta \to 0} \frac{\nabla L(\vartheta+\delta) - \nabla L(\vartheta)}{\delta}
\label{formula:hessian-matrix-estimate}
\end{equation}

\begin{equation}
\small
\eta_\vartheta^* \approx \frac{\vartheta_{t+1} - \vartheta_t}{\nabla L(\vartheta_{t+1}) -\nabla L(\vartheta_{t})}
\label{formula:adasecant-lr}
\end{equation}

\begin{equation}
\small
\vartheta_{t+1} - \vartheta_t = - \eta_t \nabla L(\vartheta_{t})
\label{formula:step-1-1}
\end{equation}

\begin{equation}
\small
\vartheta_{t+2} - \vartheta_{t+1} = - \eta_{t+1} \nabla L(\vartheta_{t+1})
\label{formula:step-1-2}
\end{equation}

\begin{equation}
\small
\eta_\vartheta^* \approx \eta_{t+1} \frac{\vartheta_{t+1} - \vartheta_t}{2\vartheta_{t+1} - \vartheta_t - \vartheta_{t+2}}
\label{formula:step-2}
\end{equation}

If the computed M-opt LR value is better than the one produced by the given LR policy to be verified, LRBench will use the computed M-opt LR values and output it as a recommended replacement. Thus, the LR verification process is performing both validation and tuning in order to produce a better LR policy for the given learning task, dataset and DNN model for a given deep learning framework (e.g., Caffe, TensorFlow, PyTorch, etc.).

\section{Experimental Analysis}\label{section:experimental-analysis}
\begin{figure*}[h!]
\centering
\scalebox{0.9}{
\subfloat[MNIST]{
  \centering
  \includegraphics[width=0.26\textwidth]{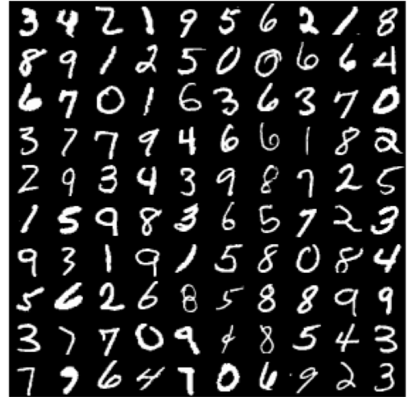}
  \label{fig:mnist-example}
} 
\subfloat[CIFAR-10]{
  \centering
  \includegraphics[width=0.255\textwidth]{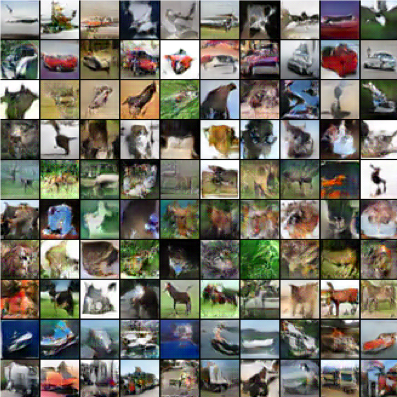}
  \label{fig:cifar10-example}
} 
\subfloat[SVHN]{
  \centering
  \includegraphics[width=0.255\textwidth]{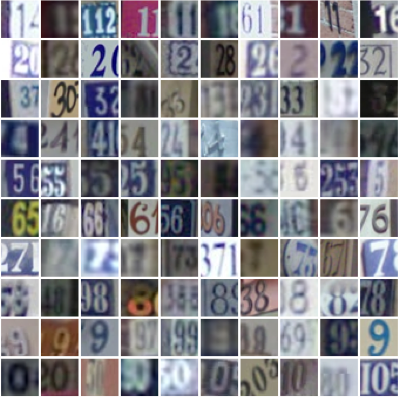}
  \label{fig:svhn-example}
} 
\subfloat[ImageNet]{
  \centering
  \includegraphics[width=0.262\textwidth]{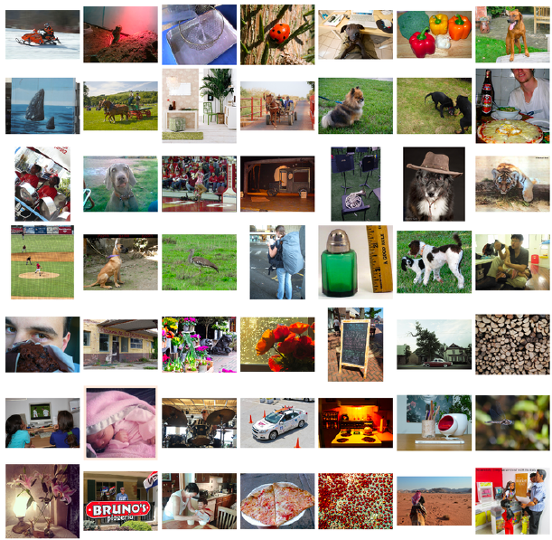}
  \label{fig:imagenet-example}
} 
} 
\caption{The Comparison of Four Datasets}
\label{fig:dataset-comparison}
\end{figure*}

We conduct experiments using LRBench on top of Caffe with four popular benchmark datasets, MNIST, CIFAR-10, SVHN and ImageNet, and three DNN models: LeNet, CNN3 and ResNet. 
Figure~\ref{fig:dataset-comparison} shows the example images from these four datasets.
We report four important results.
{\it First,} we show the impact of the LR selection on different datasets and DNN models and the experimental characterization of different types of LR policies, with respect to highest accuracy within the training time constraint.
{\it Second,} we show how to use LRBench to select and compose good LR policies for a given learning task, dataset and DNN model as well as the effectiveness of learning rate verification, and demonstrate that advanced composite LRs are beneficial for accelerating DNN training and achieving high accuracy. 
{\it Third,} we show how to efficiently choose good LR policies to achieve the desired accuracy with reduced training time cost.
{\it Fourth,} we consider other hyperparameters, such as the choice of a specific optimizer, and how different optimizers will impact on the DNN training performance with different LR policies.
In all experiments, we only vary the LR policy and keep all the other settings of hyper-parameters unchanged for the given learning task with the dataset, e.g., CIFAR-10, and the chosen DNN model, i.e., CNN3 or ResNet-32. 
We use the averaged values from 5 repeated experiments with mean$\pm$std for the most critical experiments.
All experiments reported in this paper were conducted on an Intel Xeon E5-1620 server with Nvidia GTX 1080Ti GPU, installed with Ubuntu 16.04, CUDA 8.0 and cuDNN 6.0.

\subsection{Learning Rate Policy Selection}
Here, we perform an experimental comparison of different LR policies with LRBench using MNIST and CIFAR-10 to show the impact of selecting a good LR policy on the DNN model accuracy.

\begin{table*}[h!]
\caption{Accuracy Comparison of 13 Learning Rate Policies with training LeNet on MNIST}
\label{table:lr-comparison-mnist}
\centering
\scalebox{0.9}{
\begin{tabular}{cccccccc}
\hline
Category                     & LR Function & $k (k_0)$ & $k_1$   & $\gamma$   & $p$    & $l$                            & Highest Acc (\%) \\ \hline
\multirow{4}{*}{Fixed LR}    & FIX          & 0.1    &      &         &      &                              & 11.35         \\ \cline{2-8}
                             & FIX          & 0.01   &      &         &      &                              & \textbf{99.11}         \\ \cline{2-8}
                             & FIX          & 0.001  &      &         &      &                              & 98.76         \\ \cline{2-8}
                             & FIX          & 0.0001 &      &         &      &                              & 95.10         \\ \hline
\multirow{5}{*}{Decaying LR} & NSTEP        & 0.01   &      & 0.9     &      & \footnotesize {\begin{tabular}[c]{@{}c@{}}5000,7000,\\8000,9000,9500\end{tabular}} & 99.12$\pm$0.01   \\ \cline{2-8}
                             & STEP         & 0.01   &      & 0.85    &      & 5000                         & 99.12         \\ \cline{2-8}
                             & EXP          & 0.01   &      & 0.99994 &      &                              & 99.12         \\ \cline{2-8}
                             & INV          & 0.01   &      & 0.0001  & 0.75 &                              & 99.12         \\ \cline{2-8}
                             & POLY         & 0.01   &      &         & 1.2  &                              & \textbf{99.13}         \\ \hline
\multirow{7}{*}{Cyclic LR}   & TRI          & 0.01   & 0.06 &         &      & 2000                         & 99.28$\pm$0.02   \\ \cline{2-8}
                             & TRI2         & 0.01   & 0.06 &         &      & 2000                         & 99.28$\pm$0.01   \\ \cline{2-8}
                             & TRIEXP       & 0.01   & 0.06 & 0.99994 &      & 2000                         & 99.27$\pm$0.01   \\ \cline{2-8}
                             & SIN          & 0.01   & 0.06 &         &      & 2000                         & 99.31$\pm$0.04   \\ \cline{2-8}
                             & SIN2         & 0.01   & 0.06 &         &      & 2000                         & \textbf{99.33$\pm$0.02}   \\ \cline{2-8}
                             & SINEXP       & 0.01   & 0.06 & 0.99994 &      & 2000                         & 99.28$\pm$0.04   \\ \cline{2-8}
                             & COS          & 0.01   & 0.06 &         &      & 2000                         & 99.32$\pm$0.04  \\ \hline
\end{tabular}
}
\end{table*}

\noindent{\bf MNIST (LeNet).\/}
Table~\ref{table:lr-comparison-mnist} shows the results of LeNet training on MNIST with the highest accuracy (Highest Acc (\%)) found under the Caffe default training iterations of 10,000 and batch size of 100.
We highlight three interesting observations.
{\it First,} the model accuracy trained by the baseline fixed learning rate is sensitive to the specific $k$ values. For example, when $k=0.1$, the LeNet failed to converge and resulted in a very low model accuracy, i.e., 11.35\%. Hence, it is very hard to select a proper constant LR value for the fixed learning rate.
{\it Second,} all decaying learning rates can produce high accuracy above 99.12\%. These decaying learning rate policies are recommended by LRBench based on the optimal $k$=0.01 from these fixed LRs. This observation shows that such selection of learning rate parameters by LRBench is very effective.
{\it Third,} seven CLRs recommended by LRBench produce much higher accuracy, ranging from 99.27\%$\sim$99.33\%, than other LRs. In particular, SIN2 provides the highest accuracy, that is 99.33\%, significantly outperforming the baseline fixed learning rate policies by 0.22\%$\sim$87.98\%.

This set of experiments demonstrates that LRBench can help identify good learning rate policies and significantly improve the DNN model accuracy. In general, decaying and cyclic learning rate policies can provide better training results than the baseline fixed learning rate policies. Similar observations can be found on CIFAR-10, with two different DNN models, CNN3 and ResNet-32.

\begin{table*}[h!]
\caption{Accuracy Comparison of 16 Learning Rate Policies with training CNN3 on CIFAR-10}
\label{table:lr-comparison-cifar10}
\centering
\scalebox{0.9}{
\small
\begin{tabular}{cccccccc}
\hline
Category                     & LR Function & $k (k_0)$  & $k_1$    & $\gamma$   & $p$    & $l$                   & Accuracy (\%) \\ \hline
\multirow{4}{*}{Fixed LR}    & FIX         & 0.1     &       &         &      &                     & 10.02         \\ \cline{2-8}
                             & FIX         & 0.01    &       &         &      &                     & 69.63         \\ \cline{2-8}
                             & FIX         & 0.001   &       &         &      &                     & \textbf{78.62}         \\ \cline{2-8}
                             & FIX         & 0.0001  &       &         &      &                     & 75.28         \\ \hline
\multirow{5}{*}{Decaying LR} & NSTEP       & 0.001   &       & 0.1     &      & \footnotesize{60000, 65000} & \textbf{81.61$\pm$0.18}   \\ \cline{2-8}
                             & STEP        & 0.001   &       & 0.85    &      & 5000                & 79.95         \\ \cline{2-8}
                             & EXP         & 0.001   &       & 0.99994 &      &                     & 79.28         \\ \cline{2-8}
                             & INV         & 0.001   &       & 0.0001  & 0.75 &                     & 78.87         \\ \cline{2-8}
                             & POLY        & 0.001   &       &         & 1.2  &                     & 81.51         \\ \hline
\multirow{10}{*}{Cyclic LR}  & TRI-S         & 0.001   & 0.006 &         &      & 2000                & 79.39$\pm$0.17   \\ \cline{2-8}
                             & TRI2-S        & 0.001   & 0.006 &         &      & 2000                & 79.95$\pm$0.14   \\ \cline{2-8}
                             & TRIEXP-S      & 0.001   & 0.006 & 0.99994 &      & 2000                & 80.16$\pm$0.12   \\ \cline{2-8}
                             & TRI         & 0.00005 & 0.006 &         &      & 2000                & 81.75$\pm$0.13   \\ \cline{2-8}
                             & TRI2        & 0.00005 & 0.006 &         &      & 2000                & 80.71$\pm$0.14   \\ \cline{2-8}
                             & TRIEXP      & 0.00005 & 0.006 & 0.99994 &      & 2000                & 81.92$\pm$0.13   \\ \cline{2-8}
                             & SIN         & 0.00005 & 0.006 &         &      & 2000                & 81.76$\pm$0.12   \\ \cline{2-8}
                             & SIN2        & 0.00005 & 0.006 &         &      & 2000                & 80.79$\pm$0.14   \\ \cline{2-8}
                             & SINEXP      & 0.00005 & 0.006 & 0.99994 &      & 2000                & \textbf{82.16$\pm$0.08}   \\ \cline{2-8}
                             & COS         & 0.00005 & 0.006 &         &      & 2000                & 81.43$\pm$0.14   \\ \hline
\end{tabular}
} 
\end{table*}

\noindent {\bf CIFAR-10 (CNN3).\/} Table~\ref{table:lr-comparison-cifar10} shows the empirical results of training CNN3 on CIFAR-10 with the highest Top-1 accuracy within Caffe default total training of 70,000 iterations, and the batch size is the default 100. The three triangle LR policies with S as the suffix are recommended in~\cite{clr}.
We make two interesting observations.
{\it First,} all the decaying and cyclic learning rate policies identified by LRBench outperform the best baseline fixed learning rate policy (FIX, $k$=0.001) of 78.62\% accuracy. For example, SINEXP achieved the highest accuracy and improved the accuracy of this best fixed LR by 3.54\% and the best decaying LR, NSTEP, by 0.55\%.
{\it Second,} the cyclic LRs recommended by LRBench (the last 7 LRs) achieved over 80.71\% accuracy, all outperforming the corresponding TRI-S, TRI2-S and TRIEXP-S in~\cite{clr} in terms of the highest accuracy within the training time constraint.
This demonstrates that LR selection is critical for effective DNN training. Even for the same LR function, different LR parameters will lead to different training performance.

\begin{table*}[h!]
\caption{Accuracy Comparison of 13 Learning Rate Policies with training ResNet-32 on CIFAR-10}
\label{table:lr-comparison-cifar10-resnet}
\centering
\scalebox{0.9}{
\small
\begin{tabular}{cccccccc}
\hline
Category                     & LR Function & $k (k_0)$  & $k_1$    & $\gamma$   & $p$    & $l$  & Accuracy (\%) \\ \hline
\multirow{4}{*}{Fixed LR}    & FIX         & 0.1    &     &         &      &              & \textbf{86.08}         \\
                             & FIX         & 0.01   &     &         &      &              & 85.41         \\
                             & FIX         & 0.001  &     &         &      &              & 82.66         \\
                             & FIX         & 0.0001 &     &         &      &              & 66.96         \\ \hline
\multirow{5}{*}{Decaying LR} & NSTEP       & 0.1    &     & 0.1     &      & \footnotesize{32000, 48000} & \textbf{92.38$\pm$0.04}   \\
                             & STEP        & 0.1    &     & 0.99994 &      &              & 91.10         \\
                             & EXP         & 0.1    &     & 0.85    &      & 5000         & 91.03         \\
                             & INV         & 0.1    &     & 0.0001  & 0.75 &              & 88.70         \\
                             & POLY        & 0.1    &     &         & 1.2  &              & \textbf{92.39}         \\ \hline
\multirow{7}{*}{Cyclic LR}   & TRI         & 0.0001 & 0.9 &         &      & 2000         & 76.91         \\
                             & TRI2        & 0.0001 & 0.9 &         &      & 2000         & 91.85         \\
                             & TRIEXP      & 0.0001 & 0.9 & 0.99994 &      & 2000         & \textbf{92.76$\pm$0.14}   \\
                             & SIN         & 0.0001 & 0.9 &         &      & 2000         & 72.78         \\
                             & SIN2        & 0.0001 & 0.9 &         &      & 2000         & 91.98         \\
                             & SINEXP      & 0.0001 & 0.9 & 0.99994 &      & 2000         & \textbf{92.81$\pm$0.08}   \\
                             & COS         & 0.0001 & 0.9 &         &      & 2000         & 76.77        \\ \hline       
\end{tabular}
} 
\end{table*}

\noindent {\bf CIFAR-10 (ResNet-32).\/}
ResNet can achieve over 90.00\% accuracy on CIFAR-10, a significant improvement over CNN3. Table~\ref{table:lr-comparison-cifar10-resnet} shows the results of training ResNet-32 on CIFAR-10 with the batch size 128 and the default of 64,000 training iterations.
We make two interesting observations. 
{\it First,} four LRs recommended by LRBench, i.e., NSTEP (92.38\%), POLY (92.39\%), TRIEXP (92.76\%) and SINEXP (92.81\%) achieved over 92.38\% accuracy, significantly outperforming the best baseline fixed LR with only 86.08\% accuracy by over 6.3\%.
{\it Second,} different cyclic LR policies with different cyclic LR functions may share the same LR range and step size parameters, such as TRI, TRI2, SIN, SIN2 and COS, they result in different training performance. For example, TRI2 and SIN2 achieve high accuracy over 91.85\% while the other three CLRs fail to achieve the accuracy of 77\%. This indicates that selecting a good LR policy involves both the proper LR function and proper LR parameter setting for effective DNN training.

To verify whether a given default LR policy is good for the given learning task and dataset under the chosen DNN model, LRBench will select the top $N$ policies from the LR policy database instead of performing exhaustive search. Also, for tuning and verification efficiency, we perform the experiments on $N$ machines in parallel when feasible. Consider the MNIST LeNet classifier with the NSTEP learning rate, and $N$=5, LRBench will verify this LR policy by comparing NSTEP with the top five LRs selected in the LR policy database by ranking on the accuracy for all LR policies associated with MNIST. Assuming SIN2 is one of the stored LRs in the database. Since SIN2 achieved higher accuracy (99.33\%) with LeNet, it will be chosen as one of the top five for verification. The 5 LRs can be verified in parallel by running the DNN training on MNIST, with verifying each LR policy on one machine, where SIN2 has the highest accuracy. The output of the LR verification will show the performance comparison of NSTEP (99.12\% accuracy) with the other 5 selected LR policies and also recommend SIN2 as the near-optimal replacement LR policy. Our empirical results show that the choice of $N$ is in general 3$\sim$5 for an effective LR verification with a reasonable size of LR policy database, e.g., over 100 policies.

\begin{table*}[h!]
\centering
\caption{Advanced Composite LRs for CIFAR-10 with CNN3}
\label{table:hclr-comparison-cifar10}
\small
\scalebox{0.92}{
\begin{tabular}{c|c|c}
\hline
\multicolumn{2}{c|}{LR Policy}                                                       & \multirow{2}{*}{Accuracy (\%)} \\ \cline{1-2}
Category         & LR Function                                                                 &                             \\ \hline
Decaying LR  & NSTEP ($k$=0.001, $\gamma=0.1$, $l$={[}60000,65000{]})                                                            & 81.61$\pm$0.18                                    \\ \hline
Cyclic LR    & SINEXP ($k_0$=0.00005, $k_1$=0.006, $\gamma$=0.99994, $l$=2000)                                            & 82.16$\pm$0.08                                      \\ \hline
\makecell{Advanced Composite \\ LR (Homogeneous)} & NSTEP-new ($k$=0.001, $\eta \in [0.0000008, 0.005]$)                         & 82.28                                             \\ \hline
\multirow{3}{*}{\makecell{Advanced Composite \\ LR (Heterogeneous)}} & 0-60000 iter:TRI ($k_0$=0.001, $k_1$=0.005, $l$=2000)           & \multirow{3}{*}{\textbf{82.53}}           \\ \cline{2-2}
                                  & 60000-65000 iter: TRI2 ($k_0$=0.0001, $k_1$=0.0005, $l$=1000)                        &                                                          \\ \cline{2-2}
                                  & 65000-70000 iter: TRI2 ($k_0$=0.00001, $k_1$=0.00005, $l$=500)                         &                                                          \\ \hline
\end{tabular}
} 
\end{table*}

\subsection{Multiple Learning Rate Policy Composition}
We then focus on how to compose different LR functions and parameters to form composite LRs. Two types of composite LRs should be considered, one is the homogeneous composite LR consisting of the same LR function with different LR parameters, and the other one is the heterogeneous composite LR that is comprised of different LR functions.

\begin{wrapfigure}{r}{0.45\linewidth}
    \centering
    \includegraphics[width=0.45\textwidth]{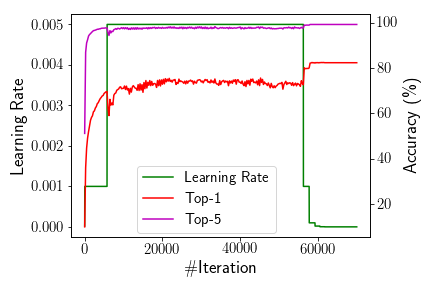}
    \caption{NSTEP-new LR (CIFAR-10, CNN3)}
    \label{fig:acc-lr-cifar10-gnstep}
\end{wrapfigure}

\noindent {\bf Homogeneous Composite Learning Rates.\/}
Recall Table~\ref{table:lr-comparison-cifar10-resnet32} and Figure~\ref{fig:lr-comparison-cifar10-resnet32}, we have shown several examples of homogeneous composite LRs. NSTEP is a special case of homogeneous composite LRs with $n$ different FIX LRs, defined by $n$ fixed LR values, denoted by $\eta_0, \eta_1, ..., \eta_{n-1}$, and $n$ training iteration boundaries $l_0, l_1, ..., l_{n-1}$, such that the LR value is $\eta_0$ when $i=0,~t< l_0$, and $\eta_i$ when $i>0$ and $l_{i-1} \le t < l_i$ for $i=1,\dots,n-1$. In the next set of experiments, we show the effectiveness of using Change-LR-On-Plateau in LRBench to select, compose and verify homogeneous composite LRs by training CNN3 on CIFAR-10 as Table~\ref{table:hclr-comparison-cifar10} and Figure~\ref{fig:acc-lr-cifar10-gnstep} show. 

By setting the initial LR value for the composite LR (NSTEP-new) at 0.001, LRBench will automatically scale it within the range of $[0.0000008, 0.005]$. The key difference between this composite LR (NSTEP-new) and the decaying LR (NSTEP) is that our NSTEP-new allows the 
increase of LR values in the training while NSTEP will only reduce the LR values. Figure~\ref{fig:acc-lr-cifar10-gnstep} shows the detailed changes of the LR value (left y-axis) and the Top-1/Top-5 accuracy (right y-axis). We observe that increasing the LR value at the beginning phase of training is beneficial, confirming that LR values should be dynamically updated to accommodate the proper learning speeds required in different training phases, instead of decaying throughout the entire training iterations.
Next, by monitoring the training loss throughout the 70,000 iterations set by the Caffe CNN3 default training time on CIFAR-10, we can identify and create a new NSTEP LR policy with improved Top-1 accuracy, compared to the decaying NSTEP for Caffe CNN3 on CIFAR-10. Also, this homogeneous composite LR (NSTEP-new) with $k=0.001$ and $\eta \in [0.0000008, 0.005]$ provides slightly improved accuracy of 82.28\% over 82.16\% of the best single policy LR SINEXP (computed over five repeated runs). It also improves the best decaying NSTEP accuracy (81.61\%) by 0.67\%.

\begin{figure*}[h!]
\centering
\subfloat[Decaying (NSTEP)]{
  \centering
  \includegraphics[width=0.3\textwidth]{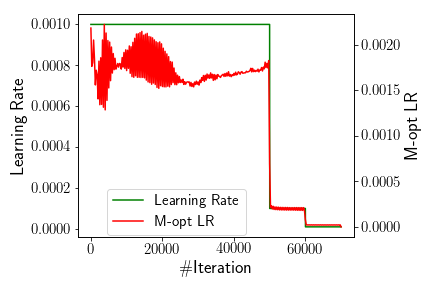}
  \label{fig:mopt-lr-nstep}
} 
\subfloat[Composite (NSTEP-new)]{
  \centering
  \includegraphics[width=0.3\textwidth]{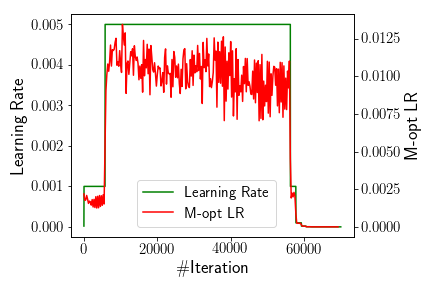}
  \label{fig:mopt-lr-gnstep}
} 
\subfloat[Composite (TRI,TRI2)]{
  \centering
  \includegraphics[width=0.3\textwidth]{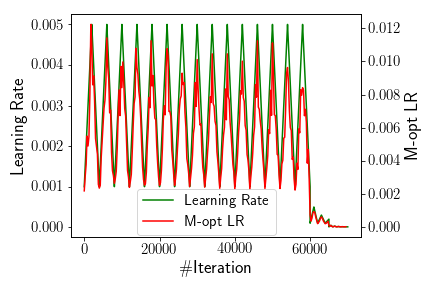}
  \label{fig:mopt-lr-hlr}
} 
\caption{M-opt Learning Rate Verification for CIFAR-10 with CNN3}
\label{fig:mopt-lr-comparison-cifar10}
\end{figure*}

\noindent {\bf Heterogeneous Composite Learning Rates.\/}
Table~\ref{table:hclr-comparison-cifar10} shows the effectiveness of a heterogeneous composite LR, which is composed by three cyclic LRs from two different CLR functions, TRI and TRI2. The highest accuracy achieved by this heterogeneous composite LR is 82.53\%, improving the best decaying NSTEP accuracy (81.61\%) by 0.92\%, and further improving the accuracy (82.28\%) of the NSTEP-new from LRBench by 0.25\%.
This shows that composite LRs offer great possibilities to select and compose good LR policies and avoid worse ones.

\noindent {\bf M-opt Learning Rate Verification.\/}
Next, we show the other method in LRBench to verify the LR values by using Formula~(\ref{formula:m-opt-lr}) to calculate the M-opt LR values every $M$ iterations. Here, we set $M$=250 and perform this set of experiments with training CNN3 for CIFAR-10 using the best decaying LR, NSTEP, and above two composite LRs, the homogeneous composite LR (NSTEP-new) and the heterogeneous composite LR with TRI and TRI2. Figure~\ref{fig:mopt-lr-comparison-cifar10} shows the results. Two interesting observations should be highlighted.
{\it First,} the LR values from the heterogeneous composite LR (TRI, TRI2) matched well with its M-opt LR values, indicating the applied LR values facilitates efficient model parameter updates. This also explains why this heterogeneous composite LR policy can achieve the highest accuracy. However, for the NSTEP and NSTEP-new LRs, the two curves show a certain interval of mismatch, which means that the LR values for this interval are not optimal.
{\it Second,} we also observe the benefits of using LRBench for selecting and composing LR policies. On the one hand, Figure~\ref{fig:mopt-lr-gnstep} further proves that the NSTEP-new dynamically tuned by LRBench is a good LR policy, demonstrating effectiveness of the tuning principles in LRBench. On the other hand, through our M-opt LR verification, practitioners can not only verify whether the chosen LR policy is optimal but also locate the potential improvements.

\begin{table*}[h!]
\centering
\caption{Learning Rate Tuning for SVHN with CNN3}
\label{table:svhn-LR-policy-comparison}
\small
\scalebox{0.92}{
\begin{tabular}{c|c|c}
\hline
\multicolumn{2}{c|}{LR Policy}  & \multirow{2}{*}{Accuracy (\%)} \\ \cline{1-2}
Category & LR Function &  \\ \hline        
Decaying LR & NSTEP ($k$=0.001, $\gamma=0.1$, $l$={{[}}60000,65000{{]}}) & 93.97 \\ \hline
Cyclic LR & SIN2 ($k_0$=0.00005, $k_1$=0.006, $\gamma$=0.99994, $l$=2000) & 19.65\\ \hline
\multirow{3}{*}{\begin{tabular}[c]{@{}c@{}}Advanced\\Composite LR\end{tabular}} & 0-60000 iter:TRI ($k_0$=0.001, $k_1$=0.005, $l$=2000) & \multirow{3}{*}{\textbf{94.23}} \\ \cline{2-2}
 & 60000-65000 iter: TRI2 ($k_0$=0.0001, $k_1$=0.0005, $l$=1000)  &  \\ \cline{2-2}
 & 65000-70000 iter: TRI2 ($k_0$=0.00001, $k_1$=0.00005, $l$=500)  &  \\ \hline
\end{tabular}
} 
\end{table*}

\noindent {\bf Learning Rate Tuning for SVHN.\/}
We further perform experiments using LRBench on another dataset SVHN. For a new dataset, it still remains a problem that whether these good LR policies are applicable to this new dataset if it shares similar features with the existing dataset in the LR policy database, such as the content and data format. We choose the Street View House Numbers (SVHN) dataset~\cite{svhn} as Figure~\ref{fig:svhn-example} shows to study this problem. SVHN consists of 73,257 training images and 26,032 testing images with 10 class labels, from 1 to 10. SVHN share more similar contents with MNIST than CIFAR-10, that is 10 digits (see Figure~\ref{fig:mnist-example}). However, SVHN and CIFAR-10 share the same data format, that is $32\times32\times3$ size for each image, and they are both colorful images (see Figure~\ref{fig:cifar10-example}). The default neural network in Caffe uses the same CNN3 architecture with 70,000 training iterations and the batch size of 100. The default LR policy for SVNH is NSTEP as shown on Table~\ref{table:svhn-LR-policy-comparison}. We use the best LR policies that produce the highest accuracy for training LeNet on MNIST (Cyclic LR: SIN2) and for training CNN3 on CIFAR-10 (Advanced Composite LR: TRI, TRI2) for this set of experiments on SVHN. We measure the highest Top-1 accuracy in Table~\ref{table:svhn-LR-policy-comparison}. Two interesting observations should be highlighted.
{\em First}, the advanced composite LR (best LR for CIFAR-10 with CNN3) achieved the highest accuracy 94.23\%, indicating that it is effective to apply the LR policy to a new dataset sharing similar features, such as the same data format and DNN architecture.
{\em Second}, even though MNIST and SVHN share the similar contents, the LR policy (SIN2) producing the highest accuracy on MNIST failed to converge the DNN model on SVHN, and it ended with only 19.65\% test accuracy.
Therefore, this set of experiments show that it is applicable to use good LR policies for a new learning tasks from empirical results, which depends on the specific similarity in the data and/or DNN model between the new dataset and the known one.

For a new dataset, LRBench can help users efficiently select and compose good LR policies. If the LR policy database contains the learning rate tuning results for an existing dataset (or DNN) that is similar to this new dataset (or DNN), LRBench will directly recommend these learning rate policies for training a DNN model on this new dataset, which avoids the typical tedious trial-and-error manual tuning.

\noindent {\bf Learning Rate Tuning for ImageNet.\/}
However, if a new dataset is very different from these existing datasets stored in the LR policy database, LRBench will perform LR tuning by performing the LR value range test and dynamic LR tuning functions to significantly narrow down the learning rate search space, such as a reduction by over 90\% (see Section~\ref{section:lr-tuning}), and quickly deliver good DNN training results, which is still more efficient than the time-consuming manual tuning process.

\begin{table*}[h!]
\centering
\caption{Learning Rate Tuning for ImageNet with ResNet-18}
\label{table:hclr-comparison-imagenet}
\small
\scalebox{0.92}{
\begin{tabular}{c|c|cc}
\hline
\multicolumn{2}{c|}{LR Policy} &
  \multirow{2}{*}{Top-1 (\%)} &
  \multirow{2}{*}{Top-5 (\%)} \\ \cline{1-2}
Category        & LR Function                                                 &       &       \\ \hline
Fixed LR    & FIX ($k$=0.1)                                                 & 50.50 & 76.51 \\ \hline
Decaying LR & STEP ($k$=0.1, $\gamma$=0.1, $l$=30 epoch)                       & 68.40 & 88.37 \\ \hline
\multirow{4}{*}{\begin{tabular}[c]{@{}c@{}}Advanced\\Composite LR\end{tabular}} &
  0-5 epoch: FIX ($k$=0.1) &
  \multirow{4}{*}{\textbf{69.05}} &
  \multirow{4}{*}{\textbf{88.76}} \\ \cline{2-2}
            & 5-35 epoch: SINEXP ($k_0$=0.1, $k_1$=0.6, $l$=15 epoch, $\gamma$=0.999987)    &       &       \\ \cline{2-2}
            & 35-50 epoch: FIX ($k$=0.01)                                       &       &       \\ \cline{2-2}
            & 50-60 epoch: FIX ($k$=0.001)                                      &       &      \\ \hline
\end{tabular}
} 
\end{table*}

For example, ImageNet is one of the large scale image datasets with over 1.2 million labeled images and 1,000 classes, which is more complex than the MNIST and CIFAR-10 datasets (see Figure~\ref{fig:imagenet-example}). It is challenging to specify proper learning rate policies for training DNNs on such a large dataset. Table~\ref{table:hclr-comparison-imagenet} shows the experimental comparison of three learning rate policies recommended by LRBench for training ResNet-18 on ImageNet for 60 epochs with the default batch size of 256. We highlight three interesting observations.
{\it First}, the baseline fixed learning rate ($k$=0.1) can train this DNN model and produce 50.50\% Top-1 and 76.51\% Top-5 accuracy.
{\it Second}, the decaying learning rate STEP recommended by LRBench can significantly improve the Top-1 and Top-5 accuracy from 50.50\% to 68.40\% by 17.9\% for Top-1 accuracy and from 76.51\% to 88.37\% by 11.86\% for Top-5 accuracy, compared to the baseline fixed LR.
{\it Third}, LRBench can further identify a good composite LR through dynamic LR tuning, which combines two LR functions, FIX and SINEXP. This advanced composite LR further improved the Top-1 and Top-5 accuracy of ResNet-18 on ImageNet to 69.05\% and 88.76\% respectively.

\begin{table*}[h!]
\caption{Cost Comparison of Learning Rate Policies for Achieving Target Accuracy}
\label{table:lr-cost-comparison}
\centering
\scalebox{0.8}{
\small
\begin{tabular}{cccc}
\hline
Dataset-Model                                                                                  & LR Function                                           & \#Iter @ Target Acc  & Speedup \\ \hline
\multirow{5}{*}{\begin{tabular}[c]{@{}c@{}}MNIST (LeNet,\\Target Acc=99.1\%)\end{tabular}}   & FIX (k=0.01)                                                       & 10000 & 1.00$\times$ \\ \cline{2-4}
& NSTEP ($k$=0.01, $\gamma$=0.9, $l$=[5000, 7000, 8000, 9000] & 10000 & 1.00$\times$ \\ \cline{2-4}
& POLY($k$=0.01, $p$=1.2)                                     & 7000  & 1.43$\times$  \\ \cline{2-4}
& SIN2 ($k_0$=0.01, $k_1$=0.06, $l$=2000)               & \textbf{3500} & 2.86$\times$ \\ \cline{2-4}
& COS ($k_0$=0.01, $k_1$=0.06, $l$=2000)                & \textbf{1500} & \textbf{6.67$\times$} \\ \hline
\multirow{4}{*}{\begin{tabular}[c]{@{}c@{}}CIFAR-10 (CNN3,\\Target Acc=80\%)\end{tabular}}  & NSTEP ($k$=0.001, $\gamma$=0.1, $l$=[60000, 65000]            & 61000 & 1.00$\times$               \\ \cline{2-4}
& POLY ($k$=0.001, $p$=1.2)                             & 59000 & 1.03$\times$               \\ \cline{2-4}
& TRIEXP ($k_0$=0.00005, $k_1$=0.006, $\gamma$=0.99994, $l$=2000)  & \textbf{16000} & 3.81$\times$ \\ \cline{2-4}
& SINEXP ($k_0$=0.00005, $k_1$=0.006, $\gamma$=0.99994, $l$=2000) & \textbf{12000} & \textbf{5.08$\times$} \\ \hline
\multirow{4}{*}{\begin{tabular}[c]{@{}c@{}}CIFAR-10 (ResNet-32,\\Target Acc=90\%)\end{tabular}} & NSTEP ($k$=0.1, $\gamma$=0.1, $l$=[32000, 48000])           & 33000 &1.00$\times$               \\ \cline{2-4}
& POLY ($k$=0.1, $p$=1.2)                             & 51000  &0.65$\times$             \\ \cline{2-4}
& TRI2 ($k_0$=0.0001, $k_1$=0.9, $l$=2000)            & \textbf{20000} &\textbf{1.65$\times$}  \\ \cline{2-4}
& SIN2 ($k_0$=0.0001, $k_1$=0.9, $l$=2000)            & \textbf{20000} & \textbf{1.65$\times$}          \\ \hline     
\end{tabular}
} 
\end{table*}

\subsection{Cost-effective Learning Rate Tuning}
In some cases, the goal for DNN training is to achieve a target accuracy threshold at a low cost, such as a smaller number of training iterations. One example is the pruning in Neural Network Search by~\cite{neuralnetworksearch}. Hence, we show the minimum \#Iterations that LRBench took to obtain the desired target accuracy for MNIST and CIFAR-10 in Table~\ref{table:lr-cost-comparison}.
We highlight two interesting observations.
{\it First,} it is very hard to achieve the target accuracy with the baseline fixed LRs. For MNIST, only one fixed LR (FIX, $k$=0.01) can achieve the target accuracy of 99.1\% with a high cost of the entire 10,000 training iterations. For CIFAR-10, all baseline fixed LRs failed to achieve the target accuracy, that is 80\% for CNN3 and 90\% for ResNet-32.
{\it Second,} cyclic LRs recommended by LRBench can achieve the desired target accuracy at a lower training cost than other LRs. For MNIST, COS reached the target accuracy of 99.10\% at 1,500 iterations, significantly reducing the training time cost by 6.67$\times$ compared to the baseline fixed LR with 10,000 training iterations. SIN2 achieves the target accuracy at 3,500 iterations, also reducing the cost by 2.86$\times$. For training CNN3 on CIFAR-10, the two CLRs, TRIEXP and SINEXP, achieved the target accuracy of 80.0\% at 12,000 and 16,000 iterations, reducing the training time by 3.81$\times$ and 5.08$\times$ respectively, compared to the baseline NSTEP of 61,000 iterations. For ResNet-32, it only took TRI2 and SIN2 20,000 iterations to reach the target accuracy of 90\% on CIFAR-10. Compared to the best decaying NSTEP of 33,000 training iterations, TRI2 and SIN2 can still significantly reduce the training cost by 1.65$\times$.

\noindent \textbf{Execution Time of LRBench.} Overall, LRBench can efficiently identify a good LR policy. If used from scratch, LRBench may take a couple of minutes to a few days to find a good LR policy, depending on the size of the training dataset and the complexity of the DNN backbone algorithm used for training.
For a small and medium dataset, such as MNIST or CIFAR-10, it takes LRBench a few minutes or hours to find a good LR policy from scratch. For MNIST, we spend about 5 minutes to obtain a good LR policy for LeNet. For CIFAR-10, both the dataset and the DNN backbone algorithm used for training are more complex, and the CIFAR-10 model training usually takes much longer time than MNIST. For example, it takes about 12 minutes to train CNN3 on CIFAR10 while it only takes about 1 minute for training LeNet on MNIST. As a result, LRBench took about 1 hour to find a good LR policy to train the CNN3 model on CIFAR-10. Moreover, if we use ResNet-32 for CIFAR-10, given that ResNet-32 takes much longer training time on CIFAR-10 (about 2 hours), LRBench also takes about $K\times 2$ hours to find a good LR policy, where $K$ is the number of candidate policies that LRBench is configured to use.
For large datasets, it takes much longer time. Using our experimental server, it takes about 10 days on ImageNet for LRBench to identify a good LR policy, due to both the large dataset and the complex neural network architecture. One way to effectively reduce such time costs is to run LRBench in parallel. Currently, users can run multiple LR policies in LRBench in parallel, each performs one of the $K$ candidate policies. We are working on an auto-parallel version of the LRBench to automate such runtime parallelism optimization.

\begin{table*}[h!]
\centering
\caption{Impact of Different Optimizers (Momentum and SGD) on training LeNet on MNIST }
\label{table:optimizer-comparison-mnist}
\small
\scalebox{0.86}{
\begin{tabular}{cccccc|cc|cc}
\hline
\multirow{2}{*}{LR} & \multirow{2}{*}{$k (k_0)$} & \multirow{2}{*}{$k_1$} & \multirow{2}{*}{$\gamma$} & \multirow{2}{*}{$p$} & \multirow{2}{*}{$l$} & \multicolumn{2}{c|}{\#Iter@Highest Acc} & \multicolumn{2}{c}{Highest Accuracy (\%)} \\ \cline{7-10} 
 &  &  &  &  &  & \redtext{Momentum} & SGD & \redtext{Momentum} & SGD \\ \hline
FIX & 0.01 &  &  &  &  & \redtext{10000} & 9500 & \redtext{99.11} & 98.71 \\ \hline
NSTEP & 0.01 &  & 0.9 &  & \scriptsize {\begin{tabular}[c]{@{}c@{}}5000,7000,\\8000,9000,9500\end{tabular}} & \redtext{10000} & 10000 & \redtext{99.12$\pm$0.01} & 98.63 \\ \hline
STEP & 0.01 &  & 0.85 &  & 5000 & \redtext{10000} & 9500 & \redtext{99.12} & 98.65 \\ \hline
EXP & 0.01 &  & 0.99994 &  &  & \redtext{10000} & 9500 & \redtext{99.12} & 98.53 \\ \hline
INV & 0.01 &  & 0.0001 & 0.75 &  & \redtext{10000} & 10000 & \redtext{99.12} & 98.58 \\ \hline
POLY & 0.01 &  &  & 1.2 &  & \redtext{7000} & 8500 & \redtext{99.13} & 98.16 \\ \hline
TRI & 0.01 & 0.06 &  &  & 2000 & \redtext{\textbf{4000} (2.5$\times$)} & 8000 & \redtext{99.28$\pm$0.02} & 99.02 \\ \hline
TRI2 & 0.01 & 0.06 &  &  & 2000 & \redtext{\textbf{4000} (2.5$\times$)} & 8000 & \redtext{99.28$\pm$0.01} & 99.03 \\ \hline
TRIEXP & 0.01 & 0.06 & 0.99994 &  & 2000 & \redtext{\textbf{4000} (2.5$\times$)} & 7500 & \redtext{99.27$\pm$0.01} & 99.05 \\ \hline
SIN & 0.01 & 0.06 &  &  & 2000 & \redtext{\textbf{4000} (2.5$\times$)} & 8000 & \redtext{99.31$\pm$0.04} & 99.03 \\ \hline
SIN2 & 0.01 & 0.06 &  &  & 2000 & \redtext{\textbf{4000} (2.5$\times$)} & 7500 & \redtext{\textbf{99.33$\pm$0.02}} & 98.92 \\ \hline
SINEXP & 0.01 & 0.06 & 0.99994 &  & 2000 & \redtext{\textbf{4000} (2.5$\times$)} & 7500 & \redtext{99.28$\pm$0.04} & 99.06 \\ \hline
COS & 0.01 & 0.06 &  &  & 2000 & \redtext{10000} & 9500 & \redtext{99.32$\pm$0.04} & \textbf{99.08} \\ \hline
\end{tabular}
} 
\end{table*}

\begin{table*}[h!]
\centering
\caption{Impact of Different Optimizers (Momentum, SGD, Adam) on training CNN3 on CIFAR-10}
\label{table:optimizer-comparison-cifar10}
\small
\scalebox{0.86}{
\begin{tabular}{cccccc|ccc|ccc}
\hline
\multirow{2}{*}{LR} & \multirow{2}{*}{$k (k_0)$} & \multirow{2}{*}{$k_1$} & \multirow{2}{*}{$\gamma$} & \multirow{2}{*}{$p$} & \multirow{2}{*}{$l$} & \multicolumn{3}{c|}{\#Iter@Highest Acc} & \multicolumn{3}{c}{Highest Accuracy (\%)} \\ \cline{7-12} 
 &  &  &  &  &  & \redtext{Momentum} & SGD & \bluetext{Adam} & \redtext{Momentum} & SGD & \bluetext{Adam} \\ \hline
FIX & 0.001 &  &  &  &  & \redtext{45000} & 70000 & \bluetext{4500} & \redtext{78.62} & 75.26 & \bluetext{69.64} \\ \hline
NSTEP & 0.001 &  & 0.1 &  & \scriptsize{60000, 65000} & \redtext{62250} & 65250 & \bluetext{7500} & \redtext{81.61$\pm$0.18} & 76.27 & \bluetext{69.98} \\ \hline
STEP & 0.001 &  & 0.85 &  & 5000 & \redtext{68000} & 69750 & \bluetext{15000} & \redtext{79.95} & 71.91 & \bluetext{69.58} \\ \hline
EXP & 0.001 &  & 0.99994 &  &  & \redtext{70000} & 65250 & \bluetext{69750} & \redtext{79.28} & 67.91 & \bluetext{70.45} \\ \hline
INV & 0.001 &  & 0.0001 & 0.75 &  & \redtext{66000} & 70000 & \bluetext{8000} & \redtext{78.87} & 70.72 & \bluetext{70.66} \\ \hline
POLY & 0.001 &  &  & 1.2 &  & \redtext{70000} & 69500 & \bluetext{0} & \redtext{81.51} & \textbf{80.84} & \bluetext{10.13} \\ \hline
TRI & 0.00005 & 0.006 &  &  & 2000 & \redtext{68000} & 68000 & \bluetext{4000} & \redtext{81.75$\pm$0.13} & 79.55 & \bluetext{69.15} \\ \hline
TRI2 & 0.00005 & 0.006 &  &  & 2000 & \redtext{70000} & 70000 & \bluetext{8000} & \redtext{80.71$\pm$0.14} & 70.93 & \bluetext{70.16} \\ \hline
TRIEXP & 0.00005 & 0.006 & 0.99994 &  & 2000 & \redtext{68000} & 67750 & \bluetext{12000} & \redtext{81.92$\pm$0.13} & 75.34 & \bluetext{\textbf{70.71}} \\ \hline
SIN & 0.00005 & 0.006 &  &  & 2000 & \redtext{68000} & 68000 & \bluetext{24000} & \redtext{81.76$\pm$0.12} & 80.01 & \bluetext{68.01} \\ \hline
SIN2 & 0.00005 & 0.006 &  &  & 2000 & \redtext{70000} & 68250 & \bluetext{8000} & \redtext{80.79$\pm$0.14} & 72.16 & \bluetext{69.38} \\ \hline
SINEXP & 0.00005 & 0.006 & 0.99994 &  & 2000 & \redtext{52000} & 64750 & \bluetext{8000} & \redtext{\textbf{82.16$\pm$0.08}} & 75.84 & \bluetext{68.99} \\ \hline
COS & 0.00005 & 0.006 &  &  & 2000 & \redtext{70000} & 70000 & \bluetext{250} & \redtext{81.43$\pm$0.14} & 79.87 & \bluetext{10.02} \\ \hline
\end{tabular}
} 
\end{table*}

\subsection{Impact of LR Policies on Different Optimizers} \label{section:impact-optimizers}
The choice of a specific DNN training optimizer is another important hyperparameter in DNN training. Given that these optimizers still require the learning rate to control the model parameter updates, we perform a set of experiments to study how different optimizers will impact on the choice of LR policies and the overall training performance on accuracy and training time. 
Table~\ref{table:optimizer-comparison-mnist} and~\ref{table:optimizer-comparison-cifar10} show the experimental results with two measurements: (1) the highest accuracy found and (2) the corresponding \#Iterations under the default training iterations of 10,000 for MNIST and 70,000 for CIFAR-10 with the default batch size set as 100.
Concretely, Table~\ref{table:optimizer-comparison-mnist} shows the choice of Momentum optimizer and SGD optimizer on different LR schedules for training LeNet MNIST models and Table~\ref{table:optimizer-comparison-cifar10} shows the choice of Momentum, SGD and Adam optimizers on different LR schedules for training CNN3 CIFAR-10 models. We make three observations:
{\it First,} the combo of the momentum optimizer and the composite LR function SIN2 offers the best performance of 99.33\% accuracy with 2.5$\times$ faster in the overall training time (10,000/4,000$=$2.5) on MNIST. 
{\it Second,} the combo of the momentum optimizer and the composite LR function of SINEXP achieves the best accuracy of 82.16\% with 1.35$\times$ faster in overall training time (70000/52000$\approx$1.35).
{\it Third,} the momentum optimizer combined with a composite LR schedule, such as SIN2 and SINEXP, can boost the overall training performance on accuracy and training time, followed by SGD optimizer. Adam optimizer has relatively lower performance no matter which LR schedule is used.

\noindent {\textbf{Summary.\/}} Through the three sets of experiments, we have shown (i) the selection of the good learning rate policies and schedules on different datasets and different DNN backbone algorithms to train a DNN model on the same dataset (CIFAR-10); (ii) the composition of multi-policy learning rate schedules for four different datasets and their corresponding learning tasks with their respective DNN training algorithms; and (iii) the cost-effective learning rate tuning with a targeted accuracy. 

LRBench can be utilized for performing semi-automated or automated learning rate tuning in several ways: 
(i)	If the user is providing a set of LR functions and/or their LR parameters, the LRBench can be used to verify whether and which of the LR policies from this given set are good or bad. 
(ii) A user can turn on the dynamic LR tuning option in LRBench. This will enable LRBench to setup the LR auto-tuning module, which monitors the training progress over a sample of training dataset in every $M$ iterations ($1\leq M \leq T/2$) during the total $T$ training iterations, and dynamically tune the learning rates by switching between the chosen candidate LR functions and the given range of each LR parameter based on the specific training progress conditions: for example, LRBench may switch to a larger learning rate to accelerate the training progress when the training stagnates on the plateau in the middle of training and switch to a smaller learning rate to facilitate the convergence of the model training when the training is close to the end of total training iterations. 
(iii) When the dynamic tuning is disabled, LRBench will verify the given candidate LR policies one by one and select the best performing LR schedule based on the accuracy and training time. 
(iv) For a new dataset and learning task, if the user does not have any specific LR policy as the initial candidate, LRBench can help such users to determine the initial LR functions and the specific LR parameters to use by leveraging the LR policy database, the LR value range test, the dataset similarity evaluation, and the M-opt LR auto-tuning method. One of our ongoing efforts is to introduce similarity evaluation methods for clustering DNN backbone algorithms into different clusters such that the DNN models with similar impact on the choice of learning rate schedules will be grouped together. 

\section{Related Work}
The learning rate is widely recognized as an important hyper-parameter for training DNN and it is critical for achieving high test accuracy~\cite{clr,understanding-lr-blog,sgdr,GTDLBenchTSC}, however, there are few studies to date dedicated to this topic. To the best of our knowledge, our work is the first to present a systematic study on the selection and composition of good LR policies. We also design and implement an LR recommendation tool, LRBench, to help practitioners to systematically select, compose and verify good LR policies that best fit their DNN training goals. Besides different LR functions that are introduced in Section~\ref{section:lr-select-compose}, we summarize the related work in the following three respects.

\noindent {\textbf{DNN Hyperparameter Search.}} The hyper-parameter search for finding the optimal settings for different hyperparameters for DNN training is one of the relevant research themes. Existing general-purpose hyperparameter search frameworks include Ray Tune~\cite{ray-tune}, Hyperopt~\cite{hyperopt}, SMAC~\cite{smac} and Optuna~\cite{optuna}. They use general hyperparameter search algorithms, e.g., grid search, random search and Bayesian optimization, and provide limited support for LR tuning, such as only supporting the fixed LR. LRBench can be used in conjunction with these hyperparameter tuning frameworks. For example, LRBench incorporates the support for Ray Tune, which allows LRBench to leverage existing hyperparameter tuning algorithms, such as grid search, random search and distributed parallel tuning functions, to further improve LR tuning efficiency.

\noindent {\textbf{Hypergradient Descent based LR Adaptation.}} The hypergradient descent based methods model the learning rate as a trainable variable and apply gradient descent to optimize an objective function w.r.t. this LR variable to perform automatic update on the LR value during DNN training~\cite{parameter-adaptation,gradient-based-hyperparameter-opt,online-lr-adaptation,lr-online-hypergradients}. However, these methods introduce an additional hyperparameter, the hypergradient LR, which still requires careful tuning.  

\noindent {\textbf{Model-update-aware LR Adaptation.}} Several orthogonal yet complementary studies focus on improving the DNN training optimizers to adaptively adjust the learning rate values for each model parameter based on the accumulated model updates, such as Adagrad~\cite{adagrad}, Adam~\cite{adam}, AdaDelta~\cite{adadelta}, and ADASECANT~\cite{adasecant}. However, these DNN optimizers still require a learning rate value to control the model parameter updates (see Section~\ref{problem-statement}). Our experiments in Section~\ref{section:impact-optimizers} demonstrate that for these optimizers, such as Adam, different learning rates still have high impacts on the DNN training efficiency and model accuracy.

\noindent \textbf{Automated Machine Learning (AutoML).} AutoML aims to minimize the dependence of manual labor and human assistance in developing high performance deep learning systems~\cite{automl-survey-state-of-the-art,automl-zero,automl-approach-tool-comparison,amc-automl-model-compression,h2o}. In addition to hyperparameter optimization~\cite{automl-survey-state-of-the-art,h2o}, recent AutoML efforts include neural architecture search~\cite{neuralnetworksearch} and AutoML for model compression~\cite{amc-automl-model-compression}.

\section{Conclusion}
We have presented a systematic approach for selecting and composing good LR policies for DNN training with three major contributions.
{\it First,} we build a systematic LR tuning mechanism to dynamically tune LR values and verify LR policies with respect to the target accuracy and/or training time constraints.
{\it Second,} we develop LRBench, an LR policy recommendation system, to facilitate efficient selection and composition of LR policies by leveraging different LR functions and parameters and to avoid bad ones for a given learning task, dataset and DNN model. Our experiments show that both homogeneous and heterogeneous learning rate compositions are attractive and advantageous for DNN training.
{\it Third,} we show that the good LR policies, including the composite LR policies, are applicable to a new dataset as well as the significant mutual impact of learning rate policies on different optimizers through extensive experiments.

\begin{acks}
This research is partially sponsored by the National Science Foundation under Grants NSF 2038029, NSF 1564097, and an IBM faculty award. Any opinions, findings, and conclusions or recommendations expressed in this material are those of the author(s) and do not necessarily reflect the views of the National Science Foundation or other funding agencies and companies mentioned above.
\end{acks}

\bibliographystyle{ACM-Reference-Format}
\bibliography{reference}


\end{document}